\def\year{2021}\relax
\documentclass[letterpaper]{article} 
\usepackage{aaai21}  
\usepackage{amsfonts}
\usepackage{amsmath}
\usepackage[titletoc]{appendix}
\usepackage{algorithm}
\usepackage{algorithmic}
\usepackage{amssymb}
\usepackage{longtable}
\usepackage{cite}
\usepackage[switch]{lineno}
\usepackage{supertabular}
\usepackage{subfigure}
\usepackage{booktabs}
\usepackage{array}
\usepackage{multirow}
\usepackage{natbib}
\usepackage{times}  
\usepackage{helvet} 
\usepackage{courier}  
\usepackage[hyphens]{url}  
\usepackage{graphicx} 
\usepackage{color}

\usepackage{graphicx}  
\usepackage{natbib}  
\usepackage{caption} 
\usepackage{verbatim}

\title{Combining Reinforcement Learning with Lin-Kernighan-Helsgaun Algorithm \\ for the Traveling Salesman Problem}

\author{
Jiongzhi Zheng \textsuperscript{\rm 1,2}, 
Kun He\textsuperscript{\rm 1}\thanks{Corresponding author.}, 
Jianrong Zhou\textsuperscript{\rm 1}, 
Yan Jin\textsuperscript{\rm 1},
Chu-Min Li \textsuperscript{\rm 1,3} \\ 
}
\affiliations{
\textsuperscript{\rm 1}School of Computer Science, Huazhong University of Science and Technology, China \\
\textsuperscript{\rm 2}Institute of Artificial Intelligence, Huazhong University of Science and Technology, China \\
\textsuperscript{\rm 3}MIS, University of Picardie Jules Verne, France\\
brooklet60@hust.edu.cn\\
}
\begin{document}
\maketitle

\begin{abstract}
We address the Traveling Salesman Problem (TSP), a famous NP-hard combinatorial optimization problem. And we propose a variable strategy reinforced approach, denoted as VSR-LKH, which combines three reinforcement learning methods (Q-learning, Sarsa and Monte Carlo) with the well-known TSP algorithm, called Lin-Kernighan-Helsgaun (LKH). VSR-LKH replaces the inflexible traversal operation in LKH, and lets the program learn to make choice at each search step by reinforcement learning. Experimental results on 111 TSP benchmarks from the TSPLIB with up to 85,900 cities demonstrate the excellent performance of the proposed method.
\end{abstract}

\section{Introduction}
Given a set of cities with certain locations, the Traveling Salesman Problem (TSP) is to find the shortest route, along which a salesman travels from a city to visit all the cities exactly once and finally returns to the starting point. An algorithm designed for TSP can also be applied to many other practical problems, such as the Vehicle Routing Problem (VRP) \cite{Bib26}, tool path optimization of Computer Numerical Control (CNC) machining \cite{Bib38}, etc. As one of the most famous NP-hard combinatorial optimization problems, TSP has become a touchstone for the algorithm design.

Numerous approaches have been proposed for solving the TSP. Traditional methods are mainly exact algorithms and heuristic algorithms, such as the exact solver Corconde $\footnote{http://www.math.uwaterloo.ca/tsp/concorde/index.html\label{Corconde}}$ and the Lin-Kernighan-Helsgaun (LKH) heuristic (\citeauthor{Bib1} \citeyear{Bib1}; \citeauthor{Bib2} \citeyear{Bib2}; \citeauthor{Bib3} \citeyear{Bib3}; \citeauthor{Bib4} \citeyear{Bib4}). With the development of artificial intelligence, there are some studies combining reinforcement learning \cite{Bib7} with (meta) heuristics to solve the TSP \cite{Bib36,Bib23}, and attaining good results 
on instances with up to 2,400 cities. More recently, Deep Reinforcement Learning (DRL) methods (\citeauthor{Bib27} \citeyear{Bib27}; \citeauthor{Bib28} \citeyear{Bib28}; \citeauthor{Bib295} \citeyear{Bib295}; \citeauthor{Bib36} \citeyear{Bib36}) have also been used to solve the TSP. However, DRL methods are hard to scale to large instances with thousands of cities, indicating that current DRL methods still have a gap to the competitive heuristic algorithms.

Heuristic algorithms are currently the most efficient and effective approaches for solving the TSP, including real-world TSPs with millions of cities.
The LKH algorithm~\cite{Bib1}, which improves the Lin-Kernighan (LK) heuristic \cite{Bib6}, is one of the most famous heuristics. LKH improves the route through the $k$-opt heuristic optimization method \cite{Bib17}, which replaces at most $k$ edges of the current tour at each search step. 
The most critical part of LKH is to make choice at each search step. In the $k$-opt process, it has to select edges to be removed and to be added. 
Differs to its predecessor LK heuristic of which each city has its own candidate set recording five (default value) nearest cities, LKH used an $\alpha$-value defined based on the minimum spanning tree \cite{Bib41} 
as a metric in selecting and sorting cities in the candidate set. The effect of candidate sets greatly improves the iteration speed and the search speed of both LK and LKH.   

However, when selecting edges to be added in the $k$-opt process, LKH traverses the candidate set of the current city in ascending order of the $\alpha$-value until the constraint conditions are met, which is inflexible and may limit its potential to find the optimal solution. 
In this work, we address the challenge of improving the LKH, and introduce a creative and distinctive method to combine reinforcement learning with the LKH heuristic. The proposed algorithm could learn to choose the appropriate edges to be added in the $k$-opt process by means of a reinforcement learning strategy.

Concretely, we first use three reinforcement learning methods, namely Q-learning, Sarsa and Monte Carlo \cite{Bib7}, to replace the inflexible traversal operation of LKH. The performance of the reinforced LKH algorithm by any one of the three methods has been greatly improved. Besides, we found that the three methods are complementary in reinforcing the LKH. For example, some TSP instances can be solved well by Q-learning, but not by Sarsa, and vice versa. Therefore, we propose a variable strategy reinforced approach, called VSR-LKH, that combines the above three algorithms to further improve the performance. The idea of variable strategy is inspired from Variable Neighborhood Search (VNS) \cite{Bib34}, which leverages the complementarity of different local search neighborhoods. 
The principal contributions 
are as follows:
\begin{itemize}
\item We propose a reinforcement learning based heuristic algorithm called VSR-LKH that significantly promotes the well-known LKH algorithm, and demonstrate its promising performance on 111 public TSP benchmarks with up to 85,900 cities.

\item We define a Q-value to replace the $\alpha$-value by combining the city distance and $\alpha$-value for the selection and sorting of candidate cities. And Q-value can be adjusted adaptively by learning from the information of many feasible solutions generated during the iterative search. 

\item Since algorithms solving an NP-hard combinatorial optimization 
problem usually need to make choice among many candidates at each search step, our approach suggests a way to improve conventional algorithms by letting them learn to make good choices intelligently.
\end{itemize}


\section{Related Work}
This section reviews related work in solving the TSP using exact algorithms, heuristic algorithms and reinforcement learning methods respectively. 

\subsubsection{Exact Algorithms.}~
The branch and bound (BnB) method is often used to exactly solve the TSP and its variants (\citeauthor{Bib8} \citeyear{Bib8}; \citeauthor{Bib9} \citeyear{Bib9}). The best-known exact solver Corconde \textsuperscript{\ref{Corconde}} is based on BnB, whose initial tour is obtained by the Chained LK algorithm \cite{Bib11}. Recently, the Corconde solver was sped up by improving the initial solution using a partition crossover \cite{Bib12}. All of these exact algorithms can yield optimal solutions, but the computational costs rise sharply along with the problem scale.


\subsubsection{Heuristic Algorithms.}~
Although heuristic algorithms cannot guarantee the optimal solution, they can obtain a sub-optimal solution within reasonable time. Heuristic algorithms for solving the TSP can be divided into three categories: tour construction algorithms, tour improvement algorithms and composite algorithms.

Each step of the tour construction algorithms determines the next city the salesman will visit until the complete TSP tour is obtained. The nearest neighbor (NN) algorithm \cite{Bib13} and ant colony algorithm \cite{Bib14} are among the most common methods for tour construction. 

The tour improvement algorithms usually make improvements on a randomly initialized tour. The $k$-opt algorithm \cite{Bib17} and genetic algorithm fall into this category. The evolutionary algorithm represented by \citet{Bib19} is one of the most successful genetic algorithms to solve TSP, which is based on an improved edge assembly crossover (EAX) operation. Its selection model can maintain the population diversity at low cost.

The composite algorithms usually use tour improvement algorithms to improve the initial solution obtained by a tour construction algorithm. The famous LKH algorithm is a composite algorithm that uses $k$-opt to improve the heuristically constructed initial tour. 

\subsubsection{Reinforcement Learning based Methods.}~ 
With the rapid development of artificial intelligence, some researchers have adopted reinforcement learning technique for solving the TSP.

One category is to combine reinforcement learning or DRL with existing (meta) heuristics. Ant-Q \cite{Bib20} and Q-ACS \cite{Bib21} replaced the pheromone in the ant colony algorithm with the Q-table in the Q-learning algorithm. However, the effect of Q-table is similar to that of the pheromone.  
\citet{Bib23} used reinforcement learning to construct mutation individuals in the successful genetic algorithm EAX-GA \cite{Bib19} and reported results on instances with up to 2,400 cities, but their proposed algorithm RMGA is inefficient compared with EAX-GA and LKH. Costa \emph{et al.} (\citeyear{Bib36}) utilized DRL algorithm to learn a 2-opt based heuristic, Wu \emph{et al.} (\citeyear{Bib500}) combined DRL method with the tour improvement approach such as 2-opt and node swap. They reported results on real-world TSP instances with up to 300 cities.

Another category is to apply DRL to directly solve the TSP. Bello \emph{et al.} (\citeyear{Bib24}) addressed TSP by using the actor-critic method to train a pointer network \cite{Bib25}. 
The S2V-DQN \cite{Bib27} used reinforcement learning to train graph neural networks so as to solve several combinatorial optimization problems, including minimum vertex cover, maximum cut and TSP. Shoma \emph{et al.} (\citeyear{Bib28}) used reinforcement learning with a convolutional neural network to solve the TSP. The graph convolutional network technique \cite{Bib35} was also applied to solve the TSP. The ECO-DQN \cite{Bib29} is an improved version of S2V-DQN, which has obtained better results than S2V-DQN on the maximum cut problem. Xing \emph{et al.} (\citeyear{Bib295}) used deep neural network combined with Monte Carlo tree search to solve the TSP.
It is generally hard for them to scale to large TSP instances with thousands cities as an effective heuristic such as LKH does. 

In this work, we aim to combine reinforcement learning with existing excellent heuristic algorithm in a more effective way, and handle efficiently large scale instances. 
We reinforce the key component of LKH, the $k$-opt, and thus promote the performance significantly. To our knowledge, this is the first work that combines reinforcement learning with the key search process of a well-known algorithm to solve the famous NP-hard TSP.

\section{The Existing LKH Algorithm}

We give a brief introduction to the LKH algorithm (see more details of LKH and its predecessor LK heuristic in Appendix). LKH uses $k$-opt \cite{Bib17} as the optimization method to improve the TSP tour. During the $k$-opt process, LKH first selects a starting point $p_1$, then alternately selects $k$ edges $\{x_1,x_2,...,x_k\}$ to be removed in the current tour and $k$ edges $\{y_1,y_2,...,y_k\}$ to be added until the stopping criterion is met ($2\leq k\leq 5$, which is not predetermined). 

\begin{figure}[htbp]
\centering
\includegraphics[width=0.4\columnwidth]{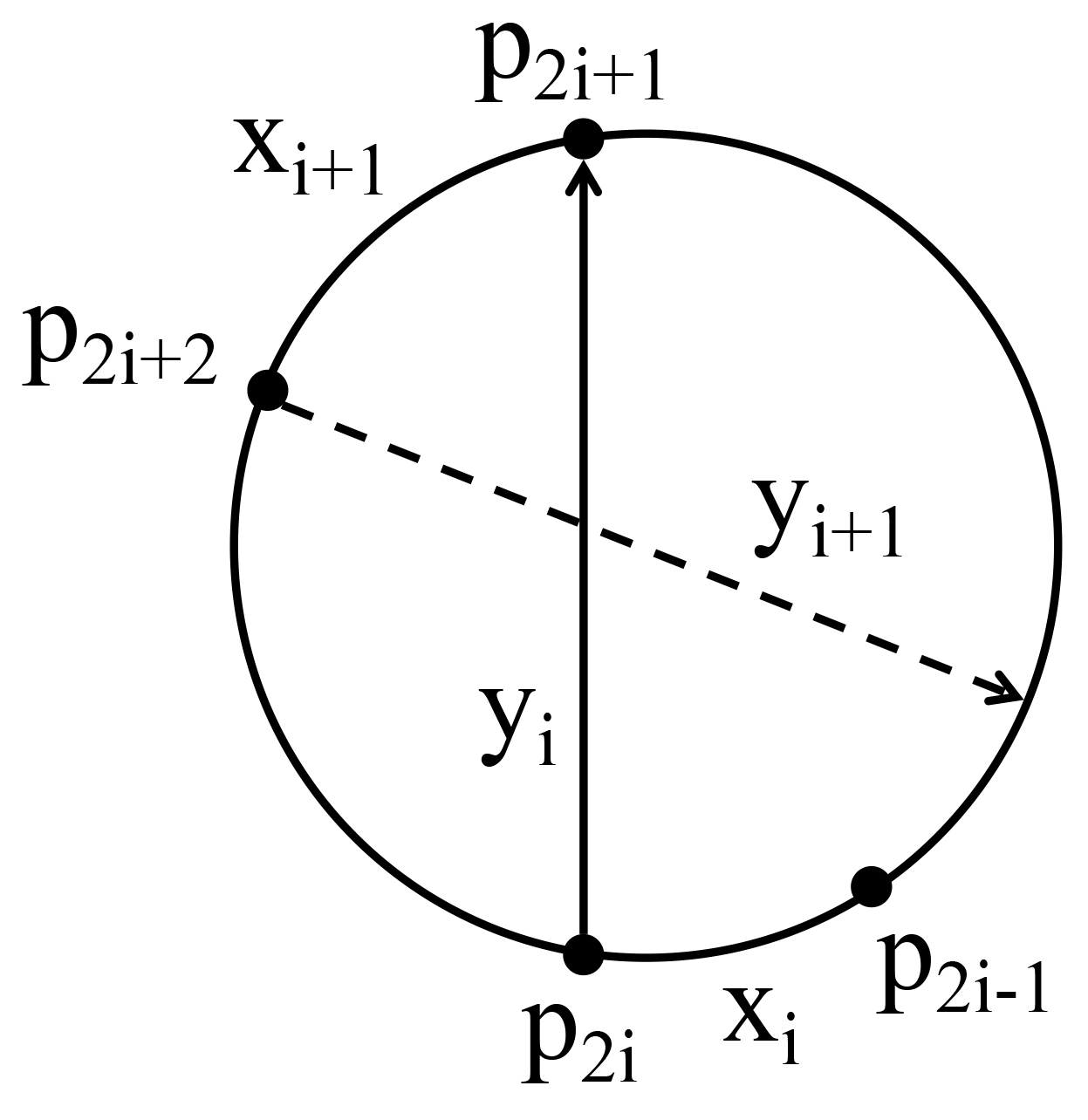} 
\caption{The choice of $x_i$, $y_i$, $x_{i+1}$ and $y_{i+1}$ in $k$-opt.}
\label{fig2}
\end{figure}

Figure \ref{fig2} shows the iterative process of $k$-opt in LKH. The key point is to select $2k$ cities $\{p_1,p_2,...,p_{2k}\}$ such that for each $i$ ($1\leq i \leq k$), $x_i=(p_{2i-1},p_{2i})$, for each $i$ ($1\leq i \leq k-1$), $y_i=(p_{2i},p_{2i+1})$ and $y_k=(p_{2k},p_1)$. And the following conditions should be satisﬁed: (1) $x_i$ and $y_i$ must share an endpoint, and so do $y_i$ and $x_{i+1}$; (2) For $i\geq2$, if $p_{2i}$ connects back to $p_1$ for $y_i$, the resulting configuration should be a tour; (3) $y_i$ is always chosen so that $\sum\nolimits_{j=1}^i(l(x_j)-l(y_j))>0$, where $l(\cdot)$ is the length of the corresponding edge; (4) Set \{$x_1, x_2, ..., x_i$\} and set \{$y_1, y_2,..., y_i$\} are disjoint. To this end, LKH first randomly selects $p_1$. After $p_{2i-1}$ is selected, $p_{2i}$ is randomly selected from the two neighbors of $p_{2i-1}$ in the current TSP tour. Then $p_{2i+1}$ is selected by traversing the candidate set of $p_{2i}$.

The current iteration of $k$-opt process will quit when a $k$-opt move is found that can improve the current TSP tour (by trying to connect $p_{2k}$ with $p_1$ as $y_k$ before selecting $p_{2k+1}$) or no edge pair, $x_i$ and $y_i$, satisfies the constraints.

\subsubsection{The $\alpha$-value.}~
The candidate set of each city in LKH stores five other cities in ascending order of the $\alpha$-values. To explain $\alpha$-value, we need to introduce the structure of 1-tree \cite{Bib41}. A 1-tree for a graph $G(V,E)$ ($V$ is the set of nodes, $E$ is the set of edges) is a spanning tree on the node set $V\backslash\{v\}$ combined with two edges from $E$ incident to node $v$, which is a special point chosen arbitrarily. A \emph{minimum} 1-tree is a 1-tree with minimum length.
Obviously, the length of the \emph{minimum} 1-tree is the lower bound of the optimal TSP solution. Suppose $L(T)$ is the length of the \emph{minimum} 1-tree of graph $G(V, E)$ and $L(T^+(i,j))$ is the length of the \emph{minimum} 1-tree required to contain edge $(i,j)$,  the $\alpha$-value of edge $(i,j)$ can be calculated by Eq. \ref{eq1}:

\begin{equation}
\alpha(i,j)=L(T^+(i,j))-L(T).
\label{eq1}
\end{equation}

\subsubsection{Penalties.}~
LKH uses a method to maximize the lower bound of the optimal TSP solution by adding \emph{penalties} \cite{Bib41}. Concretely, a $\pi$-value computed using a sub-gradient optimization method \cite{Bib30} is added to each node as a penalty when calculating the distance between two nodes:
\begin{equation}
C(i,j)=d(i,j)+\pi_i+\pi_j,
\label{eq2}
\end{equation}
where $C(i,j)$ is the cost for a salesman from city $i$ to city $j$ after adding \emph{penalties}, $d(i,j)$ is the distance between the two cities, and $\pi_i, \pi_j$ are the \emph{penalties} added to the two cities respectively. The \emph{penalties} actually change the cost matrix of the TSP. Note that this change does not change the optimal solution of the TSP, but it changes the \emph{minimum} 1-tree. Suppose $L(T_\pi)$ is the length of the \emph{minimum} 1-tree after adding the \emph{penalties}, then the lower bound $w(\pi)$ of the optimal solution can be calculated by Eq. \ref{eq3}, which is a function of set $\pi = [\pi_1, ..., \pi_n]$:
\begin{equation}
w(\pi)=L(T_\pi)-2\sum\limits_i\pi_i.
\label{eq3}
\end{equation}

The lower bound $w(\pi)$ of the optimal solution is maximized, and after adding the \emph{penalties}, the $\alpha$-value is further improved for the candidate set.

\section{The Proposed VSR-LKH Algorithm}

The proposed Variable Strategy Reinforced LKH (VSR-LKH) algorithm combines reinforcement learning with LKH. In VSR-LKH, we change the method of traversing the candidate set in the $k$-opt process and let the program automatically select appropriate edges to be added in the candidate set by reinforcement learning. We further use a variable strategy method that combines the advantages of three reinforcement learning methods, so that VSR-LKH can improve the flexibility as well as the robustness and avoid  falling into local optimal solutions. We achieve VSR-LKH on top of LKH, so VSR-LKH still retains the characteristics of LKH such as candidate set, \emph{penalties} and other improvements made by Helsgaun (\citeauthor{Bib2} \citeyear{Bib2}; \citeauthor{Bib3} \citeyear{Bib3}; \citeauthor{Bib4} \citeyear{Bib4}).

\subsection{Reinforcement Learning Framework}
Since VSR-LKH lets the program learn to make correct decisions for choosing the edges to be added, the states and actions in our reinforcement learning framework are all related to the edges to be added. And an episode corresponds to a $k$-opt process. We use the value iteration method to estimate the \emph{state-action function} $q_\pi(s,a)=\sum_{t=0}^\infty{\gamma^t{r(s_t,a_t)}}$, where $(s_t,a_t)$ is the state-action pair at time step $t$ of an episode, $s_0=s,a_0=a$, and $r$ is the corresponding reward. The detailed description of the states, actions and rewards in the reinforcement learning framework are as follows:
\begin{itemize}
\item \emph{States}: The current state of the system is a city that is going to select an edge to be added. For example in Figure \ref{fig2}, for edge $y_i$, the state corresponds to its start point $p_{2i}$. When $i=1$, the initial state $s_0$ corresponds to $p_{2i}=p_2$.
\item \emph{Actions}: For a state $s$, the action is to choose another endpoint of the edge to be added except $s$ from the candidate set of $s$. For example in Figure \ref{fig2}, for edge $y_i$, the action corresponds to its endpoint $p_{2i+1}$. When $i=1$, action $a_0$ corresponds to $p_{2i+1}=p_3$.
\item \emph{Transition}: The next state after performing the action is the next city that needs to select an edge to be added. For example, $p_{2i+2}$ is the state transferred to after executing action $p_{2i+1}$ at state $p_{2i}$. 
\item \emph{Rewards}: The reward function should be able to represent the improvement of the tour when taking an action at the current state. The reward $r(s_t,a_t)$ obtained by performing action $a_t$ at state $s_t$ can be calculated by:
\begin{equation}
r(s_t,a_t)=\begin{cases}C(a_{t-1},s_t)-C(s_t,a_t)&\text{$t>0$}\\
C(p_1,s_0)-C(s_0,a_0)&\text{$t=0$}\end{cases},
\label{eq4}
\end{equation}
where function $C(\cdot,\cdot)$ is shown in Eq. \ref{eq2}.
\end{itemize}


\subsection{Initial Estimation of the State-action Function}
We associate the estimated value of the \emph{state-action function} in VSR-LKH to each city and call it Q-value of the city. It is used as the basis for selecting and sorting the candidate cities. The initial Q-value $Q(i,j)$ for the candidate city $j$ of city $i$ is deﬁned as follows:
\begin{equation}
Q(i,j)=\frac{w(\pi)}{\alpha(i,j)+d(i,j)}.
\label{eq7}
\end{equation}

The initial Q-value defined in Eq. \ref{eq7} combines the factors of the selection and sorting of candidate cities in LK \cite{Bib6} and LKH \cite{Bib1}, which are the distance and the $\alpha$-value, respectively. Note that $\alpha$-value is based on a minimum 1-tree and is rather a global property, while the distance between two cities is a local property. Combining $\alpha$-value and distance can take the advantage of both properties. The main factor dominating the initial Q-value is the $\alpha$-value. Although the influence of the distance factor is small, it can avoid the denominator to be 0. The purpose of  $w(\pi)$ is two-folds. First, it can prevent the initial Q-value from being much smaller than the rewards. Second, it can adaptively adjust the initial Q-value for different instances. The experimental results also demonstrate that the performance of LKH can be improved by only replacing $\alpha$-value with the initial Q-value defined in Eq. \ref{eq7} to select and sort the candidate cities.



\subsection{The Reinforced Algorithms}
VSR-LKH applies reinforcement technique to learn to adjust Q-value so as to estimate the \emph{state-action function} more accurately. As the maximum value of $k$ in the $k$-opt process of LKH is as small as 5, we choose the Monte Carlo method and one-step Temporal-Difference (TD) algorithms including Q-learning and Sarsa \cite{Bib7} to improve the $k$-opt process in LKH. 

\subsubsection{Monte Carlo.}
Monte Carlo method is well-known for the model-free reinforcement learning based on averaging sample returns. In the reinforcement learning framework, there was no repetitive state or action in one or even multiple episodes. Therefore, for any state action pair $(s_t,a_t)$, the Monte Carlo method uses the episode return after taking action $a_t$ at state $s_t$ as the estimation of its Q-value. That is, 
\begin{equation}
Q(s_t,a_t)=\sum\nolimits_{i=0}^{+\infty}{r(s_{t+i},a_{t+i})}.
\label{eq8}
\end{equation}

\subsubsection{One-step TD.}
TD learning is a combination of Monte Carlo and Dynamic Programming. The TD algorithms can update the Q-values in an on-line, fully incremental fashion. In this work, we use both of the on-policy TD control (Sarsa) and off-policy TD control (Q-learning) to reinforce the LKH. The one-step Sarsa and Q-learning update the Q-value respectively as follows:
\begin{equation}
\begin{split}
Q(s_t,a_t)=&(1-\lambda)\cdot{Q(s_t,a_t)}+\\
&\lambda\cdot[r(s_t,a_t)+\gamma{Q(s_{t+1},a_{t+1})}],
\label{eq:Sarsa}
\end{split}
\end{equation}
\begin{equation}
\begin{split}
Q(s_t,a_t)=&(1-\lambda)\cdot{Q(s_t,a_t)}+\\
&\lambda\cdot[r(s_t,a_t)+\gamma{\max\limits_{a'}Q(s_{t+1},a')}],
\label{eq:QLearning}
\end{split}
\end{equation}
where $\lambda\in(0,1)$ in Eq. \ref{eq:Sarsa} and Eq. \ref{eq:QLearning} is the learning rate.

\subsection{Variable Strategy Reinforced $k$-opt}
The VSR-LKH algorithm uses the reinforcement learning technique to estimate the \emph{state-action function} to determine the optimal policy. The policy guides the program to select the appropriate edges to be added in the $k$-opt process. Moreover, a variable strategy method is added to combine the advantages of the three reinforcement learning algorithms, Q-learning, Sarsa and Monte Carlo, and leverages their complementarity. When VSR-LKH judges that the current reinforcement learning method (Q-learning, Sarsa or Monte Carlo) may not be able to continue optimizing the current TSP tour, the variable strategy mechanism will switch to another method. Algorithm 1 shows the flow of the variable strategy reinforced $k$-opt process of VSR-LKH. And the source code of VSR-LKH is available at https://github.com/JHL-HUST/VSR-LKH/.

\begin{algorithm}[!htb]
\renewcommand{\algorithmicrequire}{\textbf{Input:}}
\renewcommand{\algorithmicensure}{\textbf{Output:}}
\caption{Variable strategy reinforced $k$-opt process}
\label{alg1}
\begin{algorithmic}
\REQUIRE $\epsilon$-greedy parameter: $\epsilon$, attenuation coefficient: $\beta$, \emph{BestTour}=\emph{ChooseInitialTour}(), maximum number of iterations: \emph{MaxTrials}, variable strategy parameters: \emph{MaxNum} 
\ENSURE \emph{BestTour}
\STATE Initialize Q-values according to Eq. \ref{eq7} 
\STATE \emph{CreateCandidateSet}()
\STATE Initialize $M=1$ that corresponds to the policy (1: Q-learning, 2: Sarsa, 3: Monte Carlo), initialize $num=0$
\FOR {$i=1:MaxTrials$}
\STATE $num=num+1$, $\epsilon=\epsilon\times{\beta}$
\IF {$num\geq{MaxNum}$}
\STATE $M=M\%{3}+1,num=0$
\ENDIF
\STATE BetterTour=\emph{ChooseInitialTour}()
\REPEAT
\STATE Randomly select an unselected edge $(p_1,p_2)$ in \emph{BetterTour} as the initial edge
\STATE Initialize the set of edges to be removed $\mathcal{R}_e=\emptyset$, the set of edges to be added $\mathcal{A}_e=\emptyset$
\STATE $\mathcal{R}_e:=\mathcal{R}_e \cup \{(p_1,p_2)\}$, set $k=1$
\STATE Suppose $C_{2k}$ is the candidate set of $p_{2k}$
\WHILE {the stopping criterion is not satisfied}
\STATE $p_{2k+1}=\epsilon$-$greedy(C_{2k})$, $C_{2k}:=C_{2k}\backslash\{p_{2k+1}\}$
\IF {$p_{2k+1}$ does not satisfy the constraints}
\STATE \textbf{continue}
\ENDIF
\STATE Randomly traverse the two neighbors of $p_{2k+1}$ to select $p_{2k+2}$ that satisfies the constraints
\STATE Update $Q(p_{2k},p_{2k+1})$ according to Eq. \ref{eq8}, \ref{eq:Sarsa} or \ref{eq:QLearning} corresponding to $M$
\STATE $\mathcal{A}_e:=\mathcal{A}_e \cup \{(p_{2k},p_{2k+1})\}$
\STATE $\mathcal{R}_e:=\mathcal{R}_e \cup \{(p_{2k+1},p_{2k+2})\}$
\STATE $k=k+1$
\ENDWHILE
\STATE $\mathcal{A}_e:=\mathcal{A}_e \cup \{(p_{2k},p_1)\}$
\IF {the sum of lengths of all edges in $\mathcal{A}_e$ is less than that in $\mathcal{R}_e$}
\STATE Replace edges in $\mathcal{R}_e$ with edges in $\mathcal{A}_e$ and set the new tour as \emph{BetterTour}
\STATE \textbf{break}
\ENDIF
\UNTIL{each edge in \emph{BetterTour} has been selected as the initial edge (Note that edges $(i,j)$ and $(j,i)$ are different)}
\IF {length of \emph{BetterTour} is less than \emph{BestTour}}
\STATE Replace \emph{BestTour} with \emph{BetterTour}, $num=0$ 
\ENDIF
\STATE Exit if \emph{BestTour} is equal to the optimum (if known)
\ENDFOR
\end{algorithmic}
\end{algorithm}

In Algorithm \ref{alg1}, we apply two functions \emph{ChooseInitialTour}() and \emph{CreateCandidateSet}() in LKH to initialize the initial tour of TSP and the candidate sets of all cities. VSR-LKH uses the $\epsilon$-greedy method (\citeauthor{Bib7} \citeyear{Bib7}; \citeauthor{Bib200} \citeyear{Bib200}) to trade-off the exploration-exploitation dilemma in the reinforcement learning process. And the value of $\epsilon$ is reduced by the attenuation coefficient $\beta$ to make the algorithm inclined to exploitation as the number of iterations increases. The reinforcement learning strategy is switched if the algorithm could not improve the solution after a certain number of iterations, denoted as \emph{MaxNum}. Each iteration in the $k$-opt process of VSR-LKH needs to find cities $p_{2k+1}$ and $p_{2k+2}$ that meet the constraints before updating the Q-value. The final TSP route is stored in \emph{BestTour}. 

\section{Experimental Results}
Experimental results provide insight on why and how the proposed approach is effective, suggesting that the performance of VSR-LKH is due to the flexibility and robustness of variable strategy reinforced learning and the benefit of our Q-value definition that combines city distance and $\alpha$-value.

\subsection{Experimental Setup}
The experiments were performed on a personal computer with Intel® i5-4590 3.30 GHz 4-core CPU and 8 GB RAM. The parameters related to reinforcement learning in VSR-LKH are set as follows: $\epsilon=0.4$, $\beta=0.99$, $\lambda=0.1$, $\gamma=0.9$, the maximum iterations, \emph{MaxTrials} is equal to the number of cities in the TSP, and $MaxNum=MaxTrials/20$. Actually, VSR-LKH is not very sensitive to these parameters. The performance of VSR-LKH with different parameters is compared in Appendix. Other parameters are consistent with the example given in the LKH open source website $\footnote{http://akira.ruc.dk/\%7Ekeld/research/LKH/}$, and the LKH baseline used in our experiments is also from this website. 

All the TSP instances are from the TSPLIB $\footnote{http://comopt.ifi.uni-heidelberg.de/software/TSPLIB95/}$, including 111 symmetric TSPs. Among them, there are 77 instances with less than 1,000 cities and 34 instances with at least 1,000 cities. The number of cities in these instances ranges from 14 to 85,900. Each instance is solved 10 times by each tested algorithm. An instance is considered to be \emph{easy} if both LKH and VSR-LKH can reach the optimal solution in each of the 10 runs (for each run, the maximum number of iterations \emph{MaxTrials} equals to the number of cities), and otherwise it is considered to be \emph{hard}. All \emph{hard} instances with less than 1,000 cities include: kroB150, si175, rat195, gr229, pr299, gr431, d493, att532, si535, rat575 and gr666, a total of 11. All instances with at least 1,000 cities in TSPLIB except dsj1000, si1032, d1291, u1432, d1655, u2319, pr2392 and pla7397 are \emph{hard}, a total of 26. So there are 37 \emph{hard} instances and 74 \emph{easy} instances in TSPLIB. Note that the number in an instance name indicates the number of cities in that instance. 

We tested VSR-LKH on all the 111 symmetric TSP instances from the TSPLIB, but mainly use results on the \emph{hard} instances to compare different algorithms for clarity.

\subsection{Q-value Evaluation}
In order to demonstrate the positive influence of the Q-value we proposed in Eq. \ref{eq7} on the performance, we pick 27 \emph{hard} instances having the shortest runtime by LKH, and compare the results of three algorithms, including LKH, Q-LKH (a variant of VSR-LKH reinforced only by Q-learning, i.e. $M$ is always 1 in Algorithm \ref{alg1}) and FixQ-LKH (a variant of VSR-LKH with fixed Q-value defined by Eq. \ref{eq7} and without reinforcement learning). FixQ-LKH is equivalent to LKH but uses the initial Q-value defined by Eq. \ref{eq7} instead of $\alpha$-value to select and sort the candidate cities.


Figure \ref{fig3} shows the comparison results on TSP instances. Each instance is solved 10 times by the above three algorithms. The results are expressed by the cumulative gap on solution quality and cumulative runtime, because cumulants are more intuitive than comparing the results of each instance individually. $gap(a,j)=\frac{1}{10}\sum_{i=1}^{10}\frac{A_i-Opt_j}{Opt_j}$ is the average gap of calculating the $j$-th instance by algorithm $a$, where $A_i$ is the result of the $i$-th calculation and $Opt_j$ is the optimal solution of the $j$-th instance. The smaller the average gap, the closer the average solution is to the optimal solution. $C_{gap}(a,j)=\sum_{i=1}^{j}gap(a,i)$ is the cumulative gap. The cumulative runtime is calculated analogously. 

As shown in Figure \ref{fig3a}, the solution quality of LKH is lower than that of FixQ-LKH, indicating that LKH can be improved by only replacing $\alpha$-value with Q-value defined in Eq. \ref{eq7} to select and sort candidate cities. The results of Q-LKH compared with FixQ-LKH in Figure \ref{fig3a} demonstrate the positive effects of reinforcement learning method (Q-learning) that learns to adjust Q-value during the iterations. In addition, as shown in Figure \ref{fig3b}, there is almost no difference in efficiency among the three algorithms when solving the 27 \emph{hard} instances.


\subsection{Comparison on Reinforcement Strategies}
\begin{figure}[t]
\centering
\subfigure[Cumulative gap on the TSP instances]{
\includegraphics[width=1.0\columnwidth]{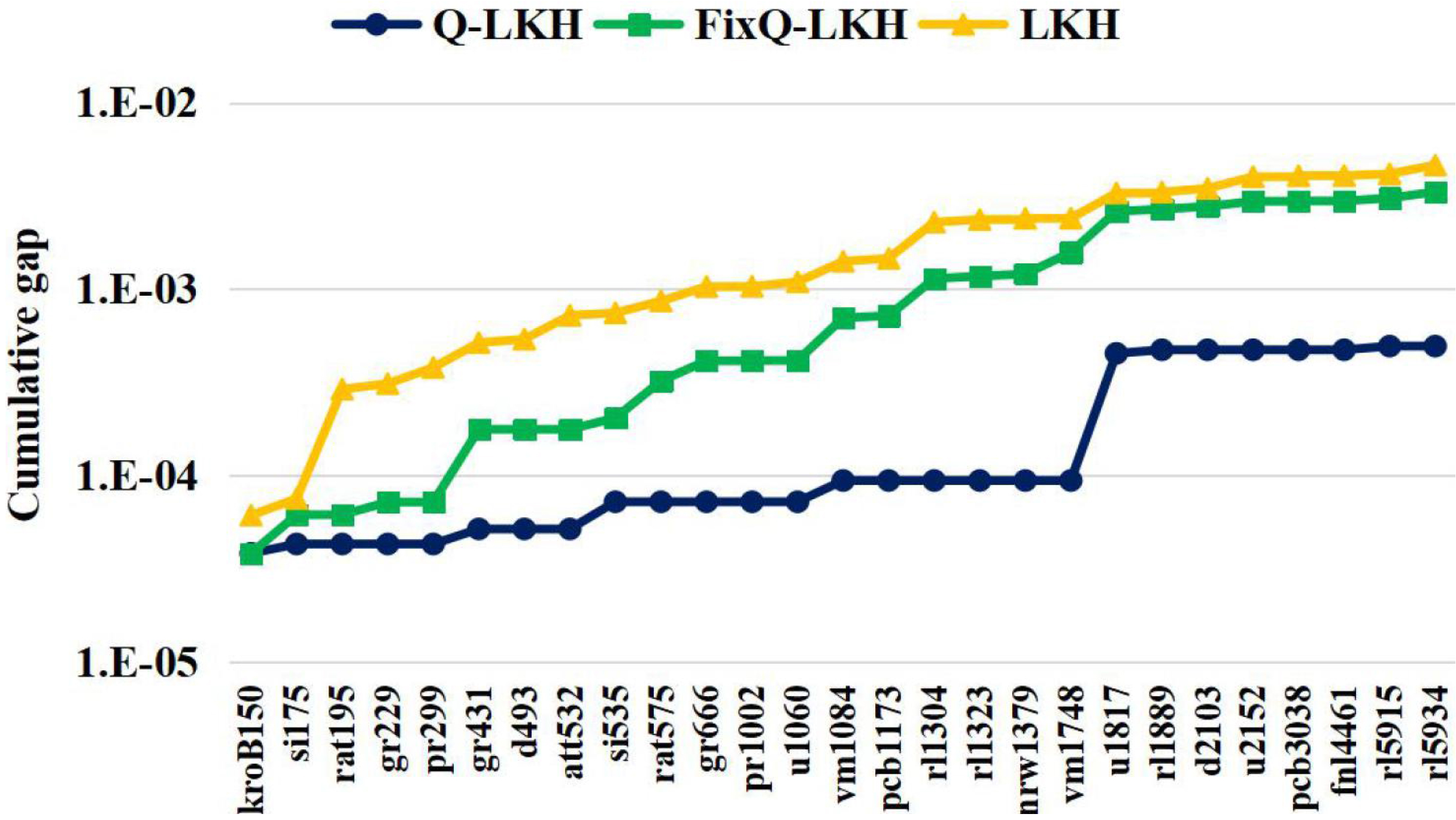} 
\label{fig3a}
}
\subfigure[Cumulative runtime on the TSP instances]{
\includegraphics[width=1.0\columnwidth]{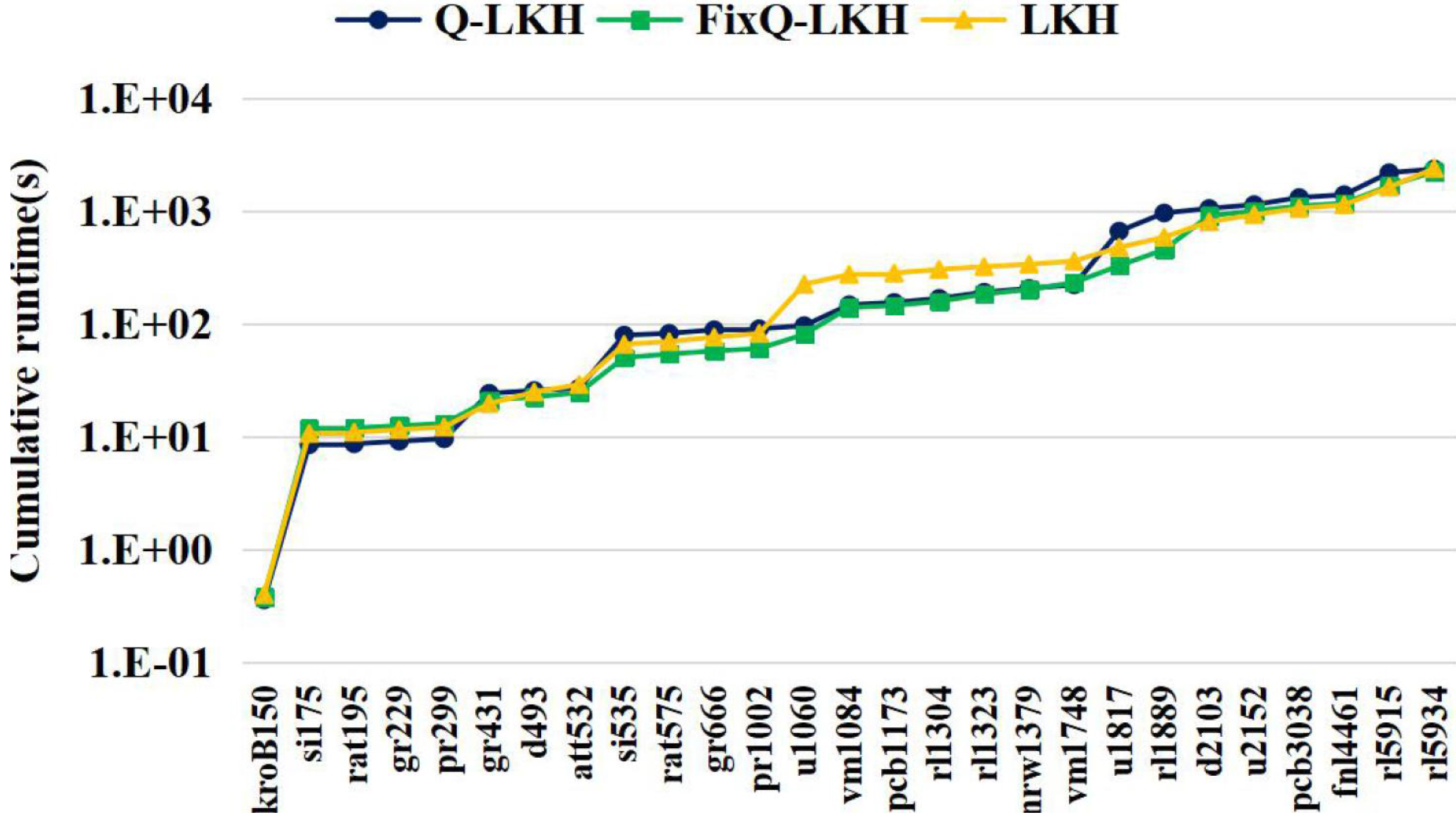} 
\label{fig3b}
}
\caption{Q-value evaluation on cumulative gap and runtime (instances are sorted in ascending order of problem size).}
\label{fig3}
\end{figure}

\begin{figure}[ht]
\centering
\subfigure[Cumulative gap on the TSP instances]{
\includegraphics[width=1.0\columnwidth]{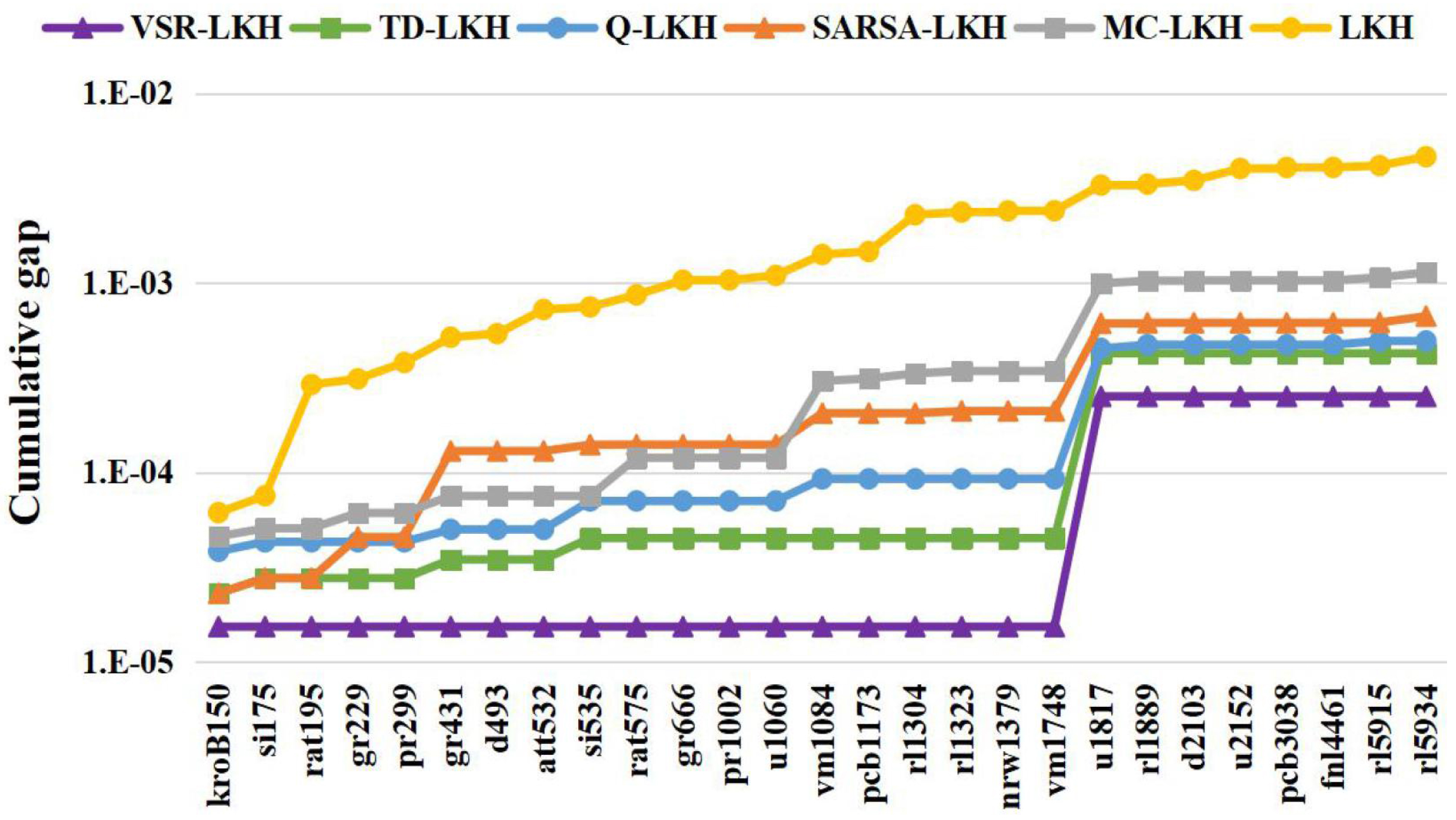} 
\label{fig4a}
}
\subfigure[Cumulative runtime on the TSP instances]{
\includegraphics[width=1.0\columnwidth]{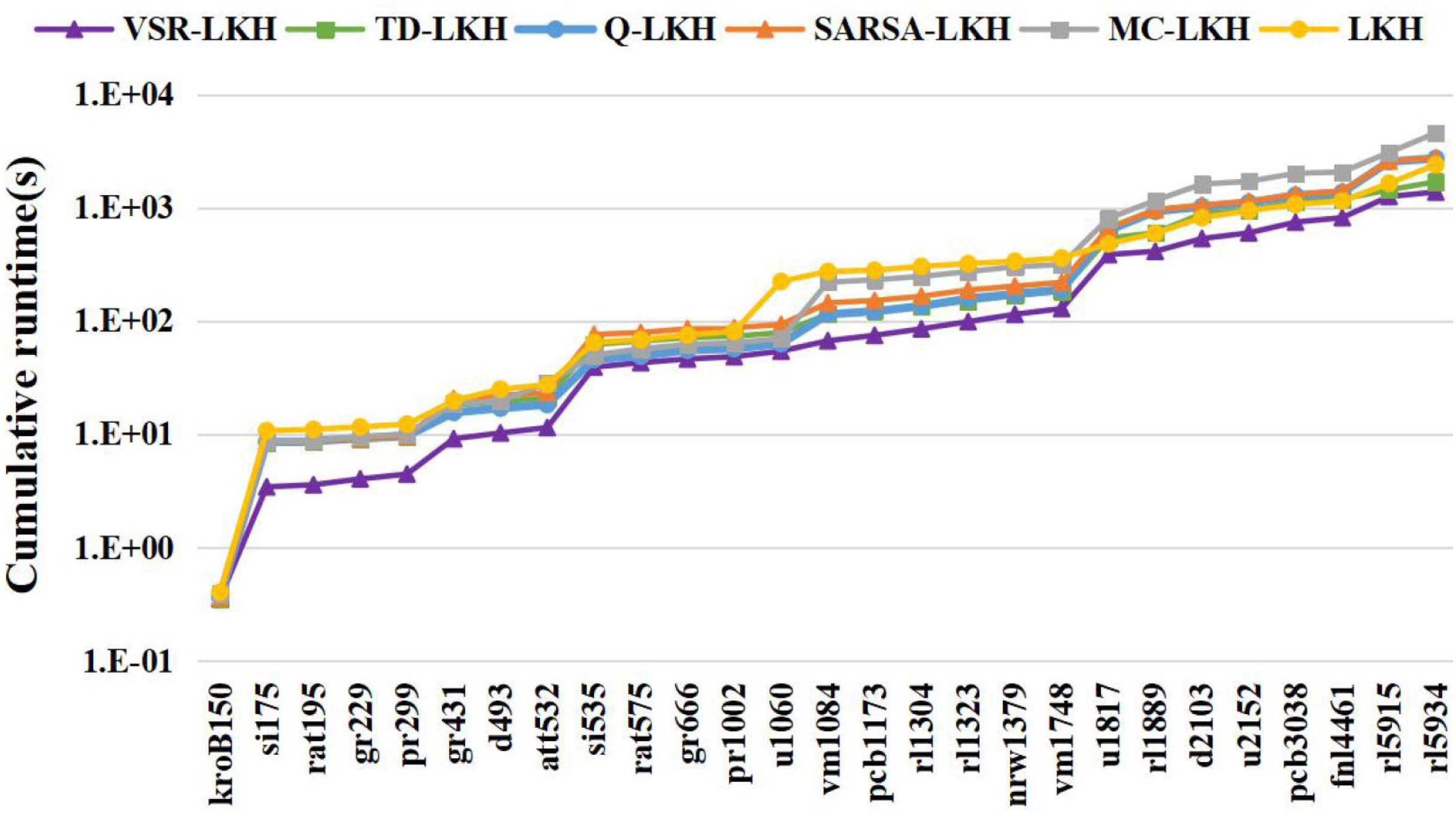} 
\label{fig4b}
}
\caption{Composite versus single reinforcement methods on cumulative gap and runtime.}
\label{fig4}
\end{figure}

Here we compare the effectiveness of the three reinforcement learning strategies (Q-learning, Sarsa and Monte Carlo) in reinforcing the LKH. The three variants of VSR-LKH reinforced only by Q-learning, Sarsa or Monte Carlo are called Q-LKH, SARSA-LKH and MC-LKH ($M$ is fixed to be always 1, 2, or 3 in Algorithm \ref{alg1} to obtain these three variants, respectively). Figure \ref{fig4} shows their results coupled with the results of LKH, VSR-LKH and TD-LKH (a variant of VSR-LKH reinforced by Q-learning and Sarsa, $M$ being 1 and 2 in turn in Algorithm \ref{alg1}) in solving the same set of 27 \emph{hard} instances in Figure \ref{fig3}. Each instance in Figure \ref{fig4} is also calculated 10 times to get the average results.

As shown in Figure \ref{fig4}, incorporating any one of the three reinforcement learning methods (Q-learning, Sarsa or Monte Carlo) to the LKH can greatly improve the performance. And the three strategies are complementary in reinforcing LKH. For example, Q-LKH has better results than SARSA-LKH or MC-LKH in most instances in Figure \ref{fig4a}, but in solving some instances like kroB150, si535 and rl5915, Q-LKH is not as good as SARSA-LKH. 

Therefore, we proposed VSR-LKH to combine the advantages of the three reinforcement methods and leverage their complementarity. Since Q-learning has the best performance in reinforcing LKH, we order Q-learning first and then Sarsa. Monte Carlo is ordered in the last since it is not good at solving large scale TSP. The order of Q-learning and Sarsa in TD-LKH is the same as in VSR-LKH.

As shown in Figure \ref{fig4}, TD-LKH and VSR-LKH further improve the solution quality compared with the three reinforced LKHs with a single method, and the two composite reinforced LKHs are significantly better than the original LKH.
In addition, though the results of MC-LKH in Figure \ref{fig4a} are the worst among single reinforcement algorithms when solving large scale TSP instances, the TD-LKH without Monte Carlo is not as efficient as VSR-LKH and is not as good as VSR-LKH in solving the instances such as gr431, si535 and u1817. Thus, adding the Monte Carlo method in VSR-LKH is effective and necessary.

Moreover, we randomly selected six instances (u574, u1060, u1817, fnl4461, rl5934 and rl11849) to compare the calculation results over the runtime of LKH and various variants of VSR-LKH. See detailed results in Appendix.

\begin{table}[b]
\center
\setlength{\tabcolsep}{1mm}{
\scalebox{0.8}{\begin{tabular}{lrrrr} \toprule
\textbf{NAME}  & \textbf{Opt.} & \textbf{VSR-LKH} & \textbf{S2V-DQN} & \textbf{Xing \emph{et al.}} \\ \hline
eil51         & 426            & \textbf{Opt.}  & 439              & 442               \\
berlin52      & 7542           & \textbf{Opt.}  &  \textbf{Opt.}    & 7598              \\
st70          & 675            & \textbf{Opt.}    & 696              & 695               \\
eil76         & 538            & \textbf{Opt.}       & 564              & 545               \\
pr76          & 108159         & \textbf{Opt.}      & 108446           & 108576            \\ 
\bottomrule
\end{tabular}}}
\caption{Comparison of VSR-LKH and DRL algorithms on five \emph{easy} instances (best results in bold).} 
\label{table1}
\end{table}

\begin{table*}[t]
\center
\scalebox{0.82}{\begin{tabular}{lrrrrrrrr} \toprule
\textbf{NAME}             & \textbf{Opt.}             & \textbf{Method} & \textbf{Best} & \textbf{Average} & \textbf{Worst} & \textbf{Success} & \textbf{Time(s)} & \textbf{Trials} \\ \hline
\multirow{2}{*}{kroB150}  & \multirow{2}{*}{26130}    & LKH             & \textbf{Opt.}          & 26131.6          & 26132          & 2/10             & 0.40              & 128.4        \\
                          &   & VSR-LKH            & \textbf{Opt.}          & 26130.4          & 26132          & 8/10             & 0.35             & 61.5            \\ \hline
\multirow{2}{*}{d493}     & \multirow{2}{*}{35002}    & LKH            & \textbf{Opt.}          & 35002.8         & 35004           & 6/10            & 5.24            & 219.6          \\
                          &   & VSR-LKH            & \textbf{Opt.}          & Opt.             & Opt.           & 10/10            & 1.14             & 10.2             \\ \hline
\multirow{2}{*}{u1060}    & \multirow{2}{*}{224094}   & LKH             & \textbf{Opt.}          & 224107.5         & 224121         & 5/10             & 142.07           & 663.3           \\
                          &   & VSR-LKH            & \textbf{Opt.}          & Opt.             & Opt.           & 10/10            & 5.68             & 13.7            \\ \hline
\multirow{2}{*}{u1817}    & \multirow{2}{*}{57201}    & LKH             & \textbf{Opt.}          & 57251.1          & 57274          & 1/10             & 116.39           & 1817.0            \\
                          &   & VSR-LKH            & \textbf{Opt.}          & 57214.5          & 57254          & 7/10             & 256.79           & 766.9           \\ \hline
\multirow{2}{*}{rl1889}   & \multirow{2}{*}{316536}   & LKH             & 316549        & 316549.8         & 316553         & 0/10             & 109.78           & 1889.0            \\
                          &   & VSR-LKH            & \textbf{Opt.}          & Opt.             & Opt.           & 10/10            & 26.30             & 91.9            \\ \hline
\multirow{2}{*}{d2103}    & \multirow{2}{*}{80450}    & LKH             & 80454         & 80462.0            & 80473          & 0/10             & 216.48           & 2103.0            \\
                          &  & VSR-LKH            & \textbf{Opt.}          & Opt.             & Opt.           & 10/10            & 121.97           & 511.8           \\ \hline
\multirow{2}{*}{rl5915}   & \multirow{2}{*}{565530}   & LKH             & 565544        & 565581.2         & 565593         & 0/10             & 494.81           & 5915.0            \\
                          &  & VSR-LKH            & \textbf{Opt.}          & Opt.             & Opt.           & 10/10            & 438.50            & 851.4           \\ \hline
\multirow{2}{*}{rl5934}   & \multirow{2}{*}{556045}   & LKH             & 556136        & 556309.8         & 556547         & 0/10             & 753.22           & 5934.0            \\
                          &  & VSR-LKH            & \textbf{Opt.}          & Opt.             & Opt.           & 10/10            & 118.78           & 144.6           \\ \hline
\multirow{2}{*}{rl11849}  & \multirow{2}{*}{923288}   & LKH             & \textbf{Opt.}          & 923362.7         & 923532         & 2/10             & 3719.35          & 10933.4         \\
                          &  & VSR-LKH            & \textbf{Opt.}          & Opt.             & Opt.           & 10/10            & 1001.11          & 751.9           \\ \hline
\multirow{2}{*}{usa13509} & \multirow{2}{*}{19982859} & LKH             & \textbf{Opt.}          & 19983103.4         & 19983569       & 1/10             & 4963.52          & 13509.0           \\
                          &  & VSR-LKH            & \textbf{Opt.}          & 19982930.2         & 19983029       & 5/10             & 25147.63    & 11900.0  \\
\bottomrule
\end{tabular}}
\caption{Comparison of VSR-LKH and LKH on some \emph{hard} instances, best results in bold (see full results in Appendix). } 
\label{table2}
\end{table*}

\subsection{Final Comparison}

We compare VSR-LKH with recent Deep Reinforcement Learning (DRL) algorithms for solving TSP, including S2V-DQN \cite{Bib27} and Xing \emph{et al.}'s method (\citeyear{Bib295}). Table \ref{table1} compares VSR-LKH and the DRL algorithms in solving five \emph{easy} instances with fewer than 100 cities that Xing \emph{et al.} picked (Xing \emph{et al.} only addressed instances with fewer than 100 cities). Column \emph{Opt.} indicates the optimal solution of the corresponding instance. Each result of VSR-LKH is the average solution of 10 runs. Table \ref{table1} shows that VSR-LKH can always yield the optimal solution and is significantly better than the DRL algorithms.

Note that DRL algorithms are hard to scale to large scale problems.
S2V-DQN provided results on 38 TSPLIB instances with the number of cities ranges from 51 to 318, but it didn't tackle larger instances due to the limitation of memory on a single graphics card. And the gap between the results of S2V-DQN and the optimal solution becomes larger when solving TSPs with more than 100 cities. 


We then compare VSR-LKH and LKH on all the 111 TSPLIB instances. Table \ref{table2} shows  detailed comparison on several typical \emph{hard} instances with various scales (see full results on all \emph{easy} and \emph{hard} instances in Appendix). The two methods both run 10 times for each TSP instance in Table \ref{table2}, and we compare the best solution, the average solution and the worst solution. Column \emph{Success} indicates the number of times the algorithm obtains optimal solution, \emph{Time} is the average calculation time of the algorithm and \emph{Trials} is the average number of iterations of VSR-LKH and LKH. 

As shown in Table \ref{table2}, 
VSR-LKH outperforms LKH on every instance, especially on four \emph{hard} instances,  rl1889, d2103, rl5915 and rl5934, that LKH did not yield optimal solution. Note that VSR-LKH also achieves good results in solving large scale TSP instances with more than 10,000 cities (rl11849 and usa13509). 
Actually, VSR-LKH can yield optimal solution on almost all the 111 instances (a total of 107, except fl1577, d18512, pla33810 and pla85900) in 10 runs. And the average solution of VSR-LKH is also \emph{Opt.} on most of the 111 instances (a total of 102, except kroB150, fl1577, u1817, usa13509, brd14051, d15112, d18512, pla33810, pla85900), indicating that VSR-LKH can always obtain optimal solution on these instances in 10 runs.

For the two super large instances, pla33810 and pla85900, we limit the maximum single runtime of LKH and VSR-LKH to 100,000 seconds due to the resource limit. And VSR-LKH can yield better results than LKH with the same runtime when solving these two instances (see Appendix).


In terms of efficiency, when solving the 111 TSPLIB instances, the average number of iterations of VSR-LKH is no more than that of LKH on 100 instances, especially for some \emph{hard} instances (1060, rl1889, rl5934 and rl11849). Although VSR-LKH takes longer time than LKH in a single iteration because it needs to trade-off the exploration and exploitation in the reinforcement learning process, it can find the optimal solution with fewer \emph{Trials} than LKH and terminate the iteration in advance. So the average runtime of VSR-LKH is less than that of LKH on most instances. In general, the proposed variable strategy reinforcement learning method greatly improves the performance of the LKH algorithm.

\section{Conclusion}
We combine reinforcement learning technique with typical heuristic search method, and propose a variable strategy reinforced method called VSR-LKH for the NP-hard traveling salesman problem (TSP). We define a Q-value as the metric for the selection and sorting of candidate cities, and change the method of traversing candidate set in selecting the edges to be added, and let the program learn to select appropriate edges to be added in the candidate set through reinforcement learning. VSR-LKH combines the advantages of three reinforcement methods, Q-learning, Sarsa and Monte Carlo, and further improves the flexibility and robustness of the proposed algorithm. 
Extensive experiments on public benchmarks show that VSR-LKH outperforms signiﬁcantly the famous heuristic algorithm LKH. 


VSR-LKH is essentially a reinforcement on the $k$-opt process of LKH. Thus, other algorithms based on $k$-opt could also be strengthened by our method. 
Furthermore, our work demonstrates the feasibility and privilege of incorporating reinforcement techniques with heuristics in solving classic combinatorial optimization problems. 
In future work, we plan to apply our approach in solving 
the constrained TSP and the vehicle routing problem.



\section{Acknowledgments}
This work is supported by National Natural Science Foundation (62076105).

\bibliographystyle{aaai}
\bibliography{VSR-LKH}

\newpage

\twocolumn[
\begin{@twocolumnfalse}
\section*{\centering{\LARGE{Appendix}}}
~\\
\end{@twocolumnfalse}
]


In the Appendix, we provide detailed description of the existing LKH algorithms, and more experimental results.

\section{The Existing LKH Algorithms}
In this section, we first introduce the Lin-Kernighan (LK) heuristic \cite{Bib6}, which is the predecessor algorithm of LKH, then explain the main improvements of LKH over LK. 

\subsection{The Lin-Kernighan Algorithm}
\begin{figure}[ht]
\centering
\includegraphics[width=0.8\columnwidth]{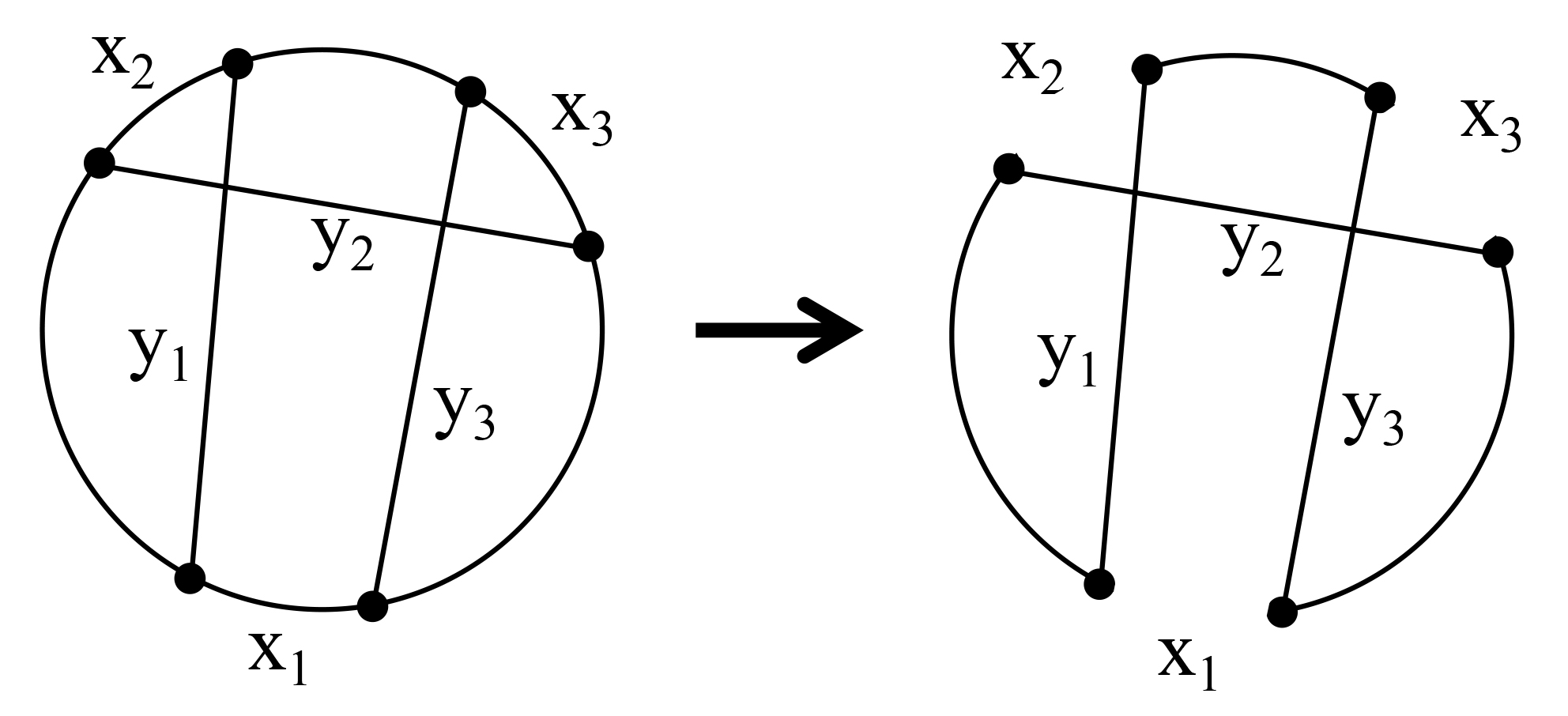} 
\caption{An illustration of 3-opt move.}
\label{figA1}
\end{figure}

LK heuristic uses $k$-opt \cite{Bib17} as the optimization method to find optimal or sub-optimal solutions. Figure \ref{figA1} illustrates a 3-opt move. The 3-opt process runs iteratively. At each iteration it attempts to change the tour by cutting three edges $x_1$, $x_2$, $x_3$ and replacing with three edges $y_1$, $y_2$, $y_3$, then picks a better tour from the two. The key point of $k$-opt is how to select these $x_i$ and $y_i$, $i\in\{1, 2,..., k\}$, aka the edges to be removed in current TSP tour and the edges to be added.

The LK algorithm has two main contributions. Firstly, LK relieves the limitation that traditional $k$-opt algorithm \cite{Bib17} needs to determine the value of $k$, and thus improves the flexibility. Secondly, LK uses the candidate set to help the $k$-opt process to select the edges to be added. The LK algorithm assumes that an edge with smaller distance is more likely to appear in the optimal solution, and the candidate set of each city stores five nearest cities in ascending order of the city distance. 

\begin{figure}
\centering
\includegraphics[width=0.4\columnwidth]{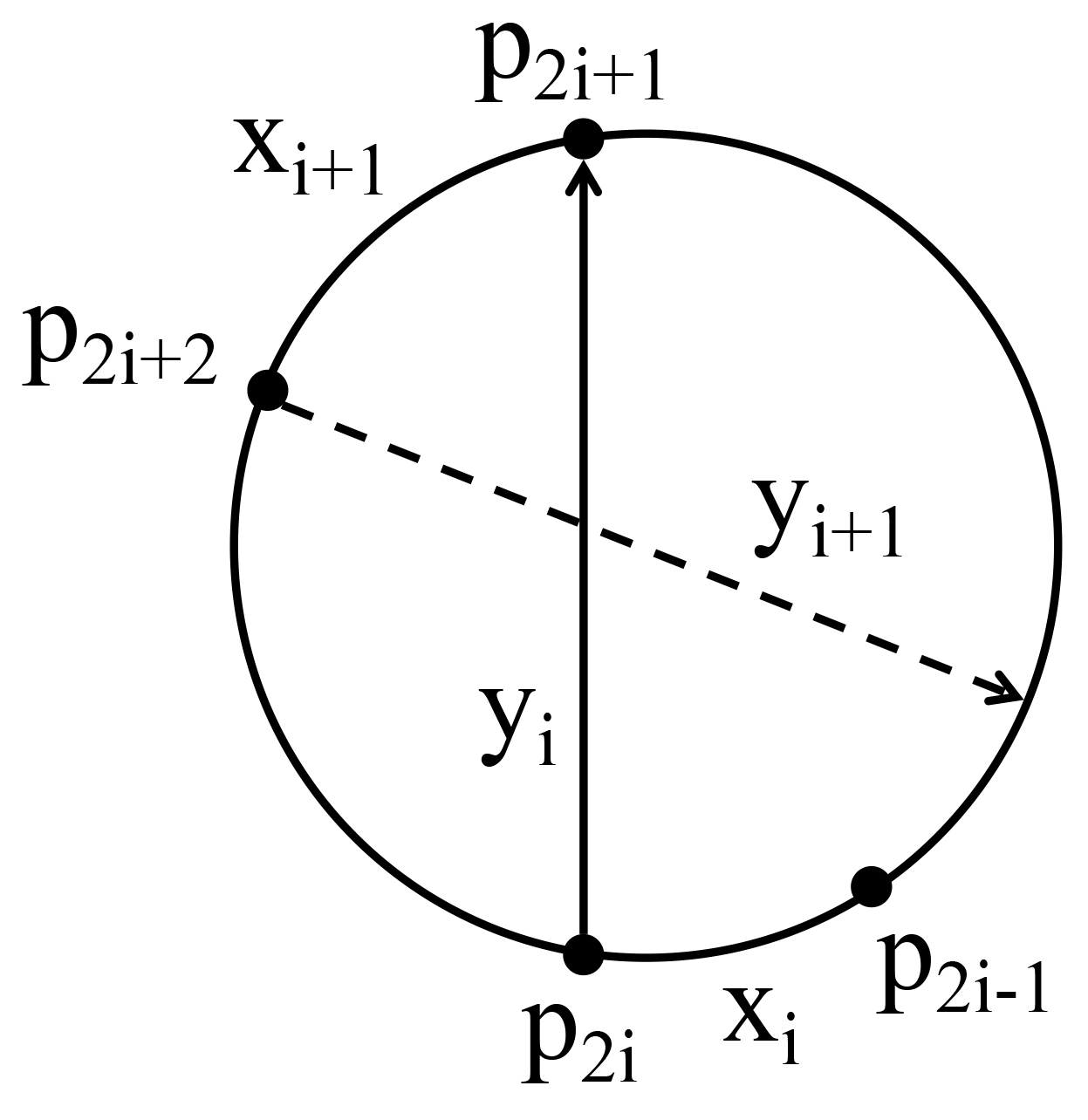} 
\caption{The choice of $x_i$, $y_i$, $x_{i+1}$ and $y_{i+1}$.}
\label{fig22}
\end{figure}

Figure \ref{fig22} illustrates the iterative process of $k$-opt, that is, the process of selecting the edges to be removed and the edges to be added. Suppose an initial tour has been obtained by a heuristic method, LK first randomly selects a point $p_1$ and let $i=1$, then the edges to be removed and the edges to be added are selected alternately. When iterating to $p_{2i-1}$, the edge $x_i$ needs to be selected. In fact, LK regards the selection of $x_i$ as the selection of $p_{2i}$. The method of selecting $p_{2i}$ is to randomly select a point connected with $p_{2i-1}$ in the current TSP tour. Similarly, the selection of $y_i$ is regarded as the selection of $p_{2i+1}$. The method for the algorithm to select $p_{2i+1}$ is to sequentially traverse the candidate set of $p_{2i}$ until it finds a city to be $p_{2i+1}$ that meets the constraints. The constraints of selecting $p_{2i}$ and $p_{2i+1}$ in LK are as follows:

\begin{itemize}
\item $x_i$ and $y_i$ must share an endpoint, and so do $y_i$ and $x_{i+1}$.
\item For $i\geq2$, if $p_{2i}$ connects back to $p_1$ for $y_i$, the resulting configuration should be a tour.
\item $y_i$ is always chosen so that $g_i=\sum\limits_{j=1}^i(l(x_j)-l(y_j))>0$, where $g_i$ is the gain of replacing \{$x_1, x_2, ..., x_i$\} with \{$y_1, y_2, ..., y_i$\}. $l(\cdot)$ is the length of the corresponding edge.
\item Set \{$x_1, x_2, ..., x_i$\} and set \{$y_1, y_2, ..., y_i$\} are disjoint.
\end{itemize}

Let $g^*$ denote the best improvement in the current $k$-opt process, and its initial value is set to 0. When $i\geq2$, LK sets $y'=(p_{2i},p_1)$ and calculates $g'=g_{i-1}+l(x_i)-l(y')$, and sets $g^*=g'$ if $g'>g^*$. The iteration of $k$-opt process in LK will stop when no link of $x_i$ and $y_i$ satisfy the above constraints or $g_i\leq{g^*}$. When the iteration stops, if $g^*>0$, then there is a $k$-opt move that can improve the current TSP route. The whole algorithm will stop when the optimal solution of TSP is found or the maximum number of iterations is reached.

\subsection{Main Improvements of LKH}
\subsubsection{The $\alpha$-measure}
LKH \cite{Bib1} proposed an $\alpha$-value instead of distance as the measure for selecting and sorting cities in the candidate set, called $\alpha$-measure, to improve the quality of the candidate set.

\begin{figure}[h]
\centering
\includegraphics[width=0.5\columnwidth]{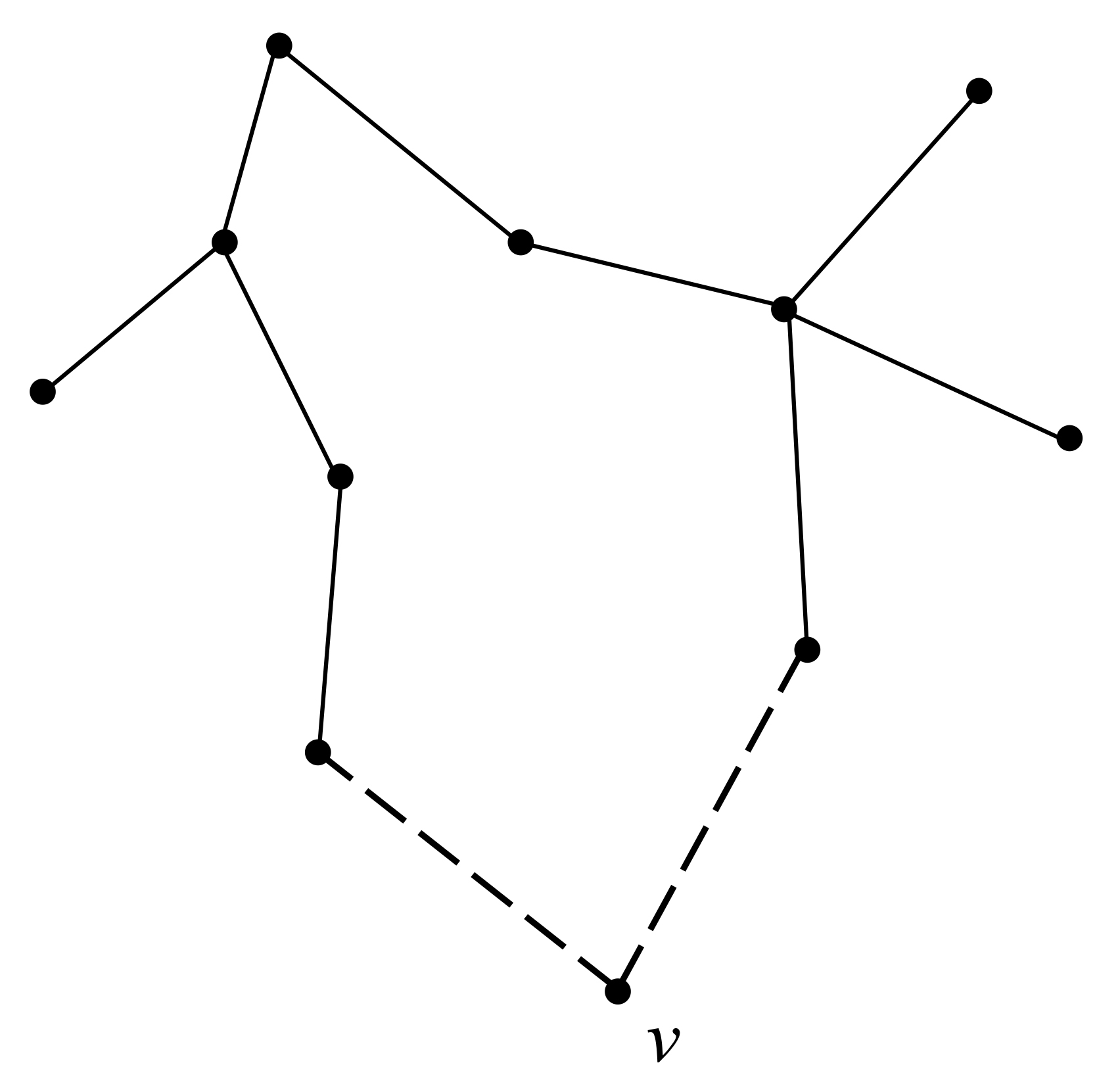} 
\caption{An example of 1-tree.}
\label{fig55}
\end{figure}

To explain the concept of $\alpha$-value, we need to introduce the structure of 1-tree. A 1-tree for a graph $G(V, E)$ ($V$ is the set of nodes, $E$ is the set of edges) is a spanning tree on the node set $V\backslash\{v\}$ combined with two edges from $E$ incident to a node $v$, which is a special point chosen arbitrarily. Figure \ref{fig55} shows a 1-tree. A \emph{minimum} 1-tree is a 1-tree with minimum length, which can be found by determining the minimum spanning tree on $G$ and then adding an edge corresponding to the second nearest neighbor of one of the leaves of the tree. The leaf chosen is the one that has the longest second nearest neighbor distance. Obviously, the length of the \emph{minimum} 1-tree is the lower bound of the optimal TSP solution. An important observation is that 70-80\% of the edges in the optimal solution of TSP are also in the \emph{minimum} 1-tree \cite{Bib1}. Therefore, the LKH algorithm tries to define an $\alpha$-value using the \emph{minimum} 1-tree to replace the distance used by LK. The equation of calculating the $\alpha$-value of an edge $(i,j)$ is as follows:
\begin{equation}
\alpha(i,j)=L(T^+(i,j))-L(T),
\label{eq:alpha}
\end{equation}
where $L(T)$ is the length of the \emph{minimum} 1-tree of graph $G(V, E)$ and $L(T^+(i,j))$ is the length of the \emph{minimum} 1-tree required to contain edge $(i,j)$. That is, the $\alpha$-value of an edge is equal to the length increased if the \emph{minimum} 1-tree needs to include this edge. The LKH algorithm considers that an edge with smaller $\alpha$-value is more likely to appear in the optimal solution. This $\alpha$-measure has greatly improved the algorithm performance.

\subsubsection{Penalties}
In order to further enhance the advantage of the candidate set, LKH uses a method to maximize the lower bound of the optimal TSP solution by adding \emph{penalties} \cite{Bib41}. Concretely, a $\pi$-value computed using a sub-gradient optimization method \cite{Bib30} is added to each node as \emph{penalties} when calculating the distance between two nodes:
\begin{equation}
C(i,j)=d(i,j)+\pi_i+\pi_j,
\label{eq22}
\end{equation}
where $C(i,j)$ is the cost for a salesman from city $i$ to city $j$ after adding \emph{penalties}, $d(i,j)$ is the distance between the two cities, and $\pi_i, \pi_j$ are the \emph{penalties} added to the two cities respectively. The \emph{penalties} actually change the cost matrix of TSP. Note that this change does not change the optimal solution of the TSP, but it changes the \emph{minimum} 1-tree. Suppose $L(T_\pi)$ is the length of the \emph{minimum} 1-tree after adding the \emph{penalties}, then the lower bound $w(\pi)$ of the optimal solution can be calculated by Eq. \ref{eq33}, which is a function of set $\pi = [\pi_1, ..., \pi_n]$:
\begin{equation}
w(\pi)=L(T_\pi)-2\sum\limits_i\pi_i.
\label{eq33}
\end{equation}

The lower bound $w(\pi)$ of the optimal solution is maximized, and after adding the \emph{penalties}, the $\alpha$-value is further improved for the candidate set. 

In general, LKH \cite{Bib1} uses $\alpha$-measure and \emph{penalties} to improve the performance of LK. Moreover, the iteration of $k$-opt process in LKH will stop immediately if a $k$-opt move is found that can improve the current TSP tour. And LKH uses a 5-opt move as the basic move which means the maximum value of $k$ in the $k$-opt process of LKH is 5.

\section{More Experimental Results}
In this section, we provide more experimental results, including the evaluation on the parameter sentivity of our algorithm, detailed performance comparision of VSR-LKH variants and LKH over the runtime, and the full results on all the 111 TSP benchmark instances.

\subsection{Evaluation on Parameter Sensitivity}
To evaluate the parameter sensitive as well as the robustness of VSR-LKH and choose appropriate reinforcement learning parameters, we do ablation study using different parameters include $\epsilon$, $\beta$, $\lambda$, $\gamma$ and \emph{MaxNum}. Note that when comparing each of the parameters, the others are fixed to default values ($\epsilon=0.4$, $\beta=0.99$, $\lambda=0.1$, $\gamma=0.9$ and $MaxNum=n/20$, $n$ is the number of the cities). 

\begin{figure}[t]
\centering
\subfigure[The results with different $\epsilon$]{
\includegraphics[width=0.3\columnwidth]{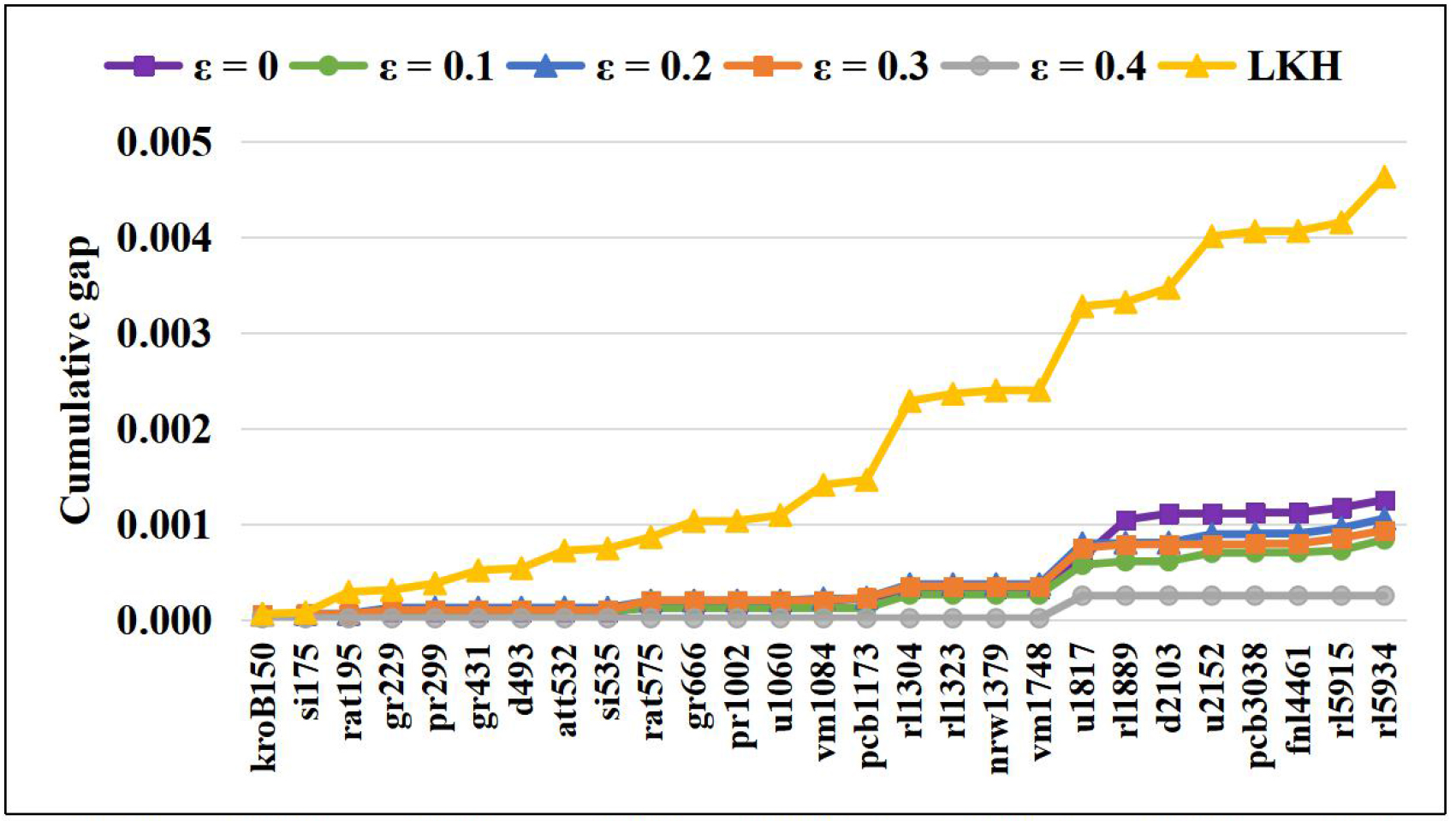} 
\label{fig:pa}
}
\subfigure[The results with different $\beta$]{
\includegraphics[width=0.3\columnwidth]{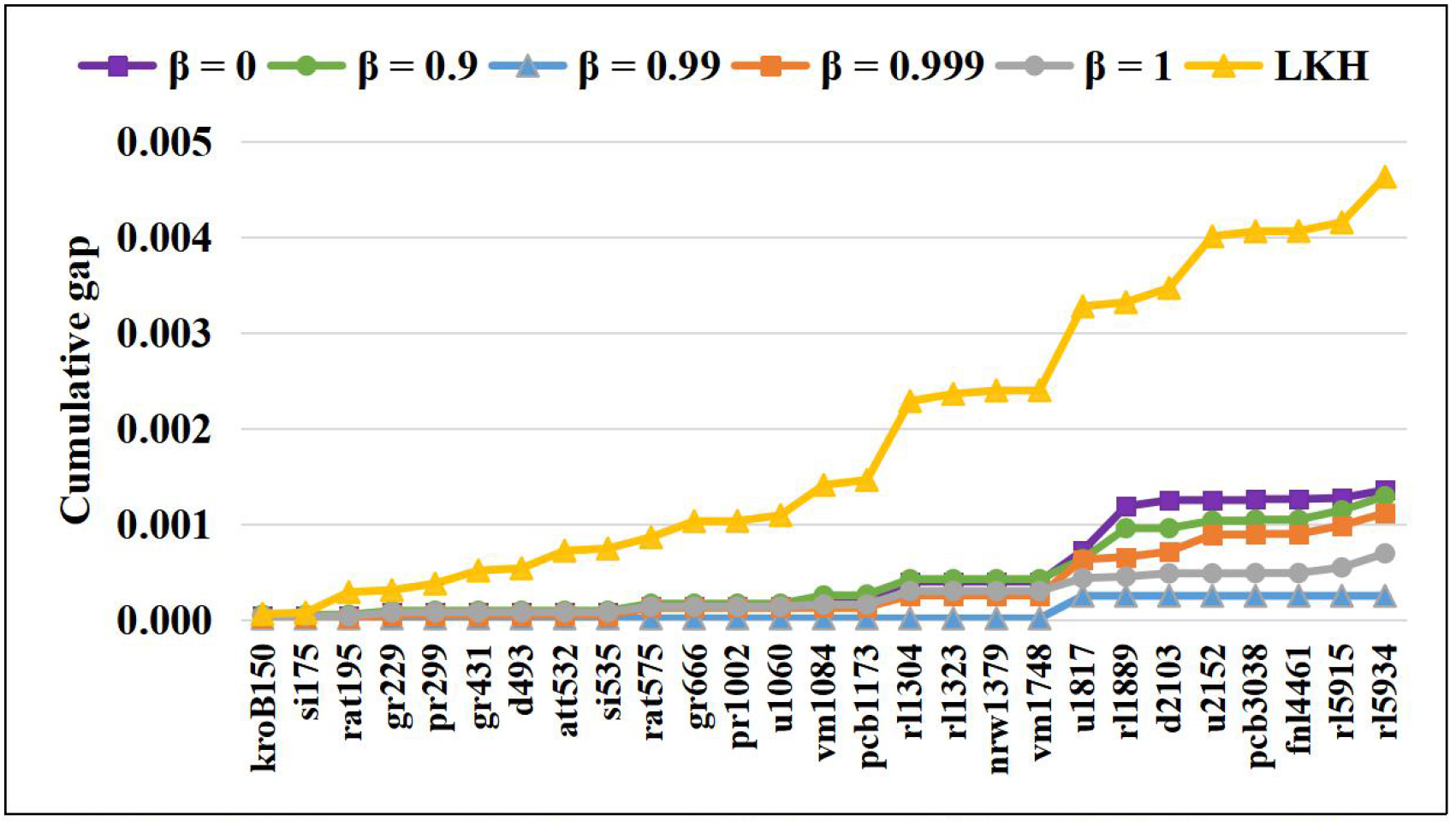} 
\label{fig:pb}
}
\subfigure[The results with different $\lambda$]{
\includegraphics[width=0.31\columnwidth]{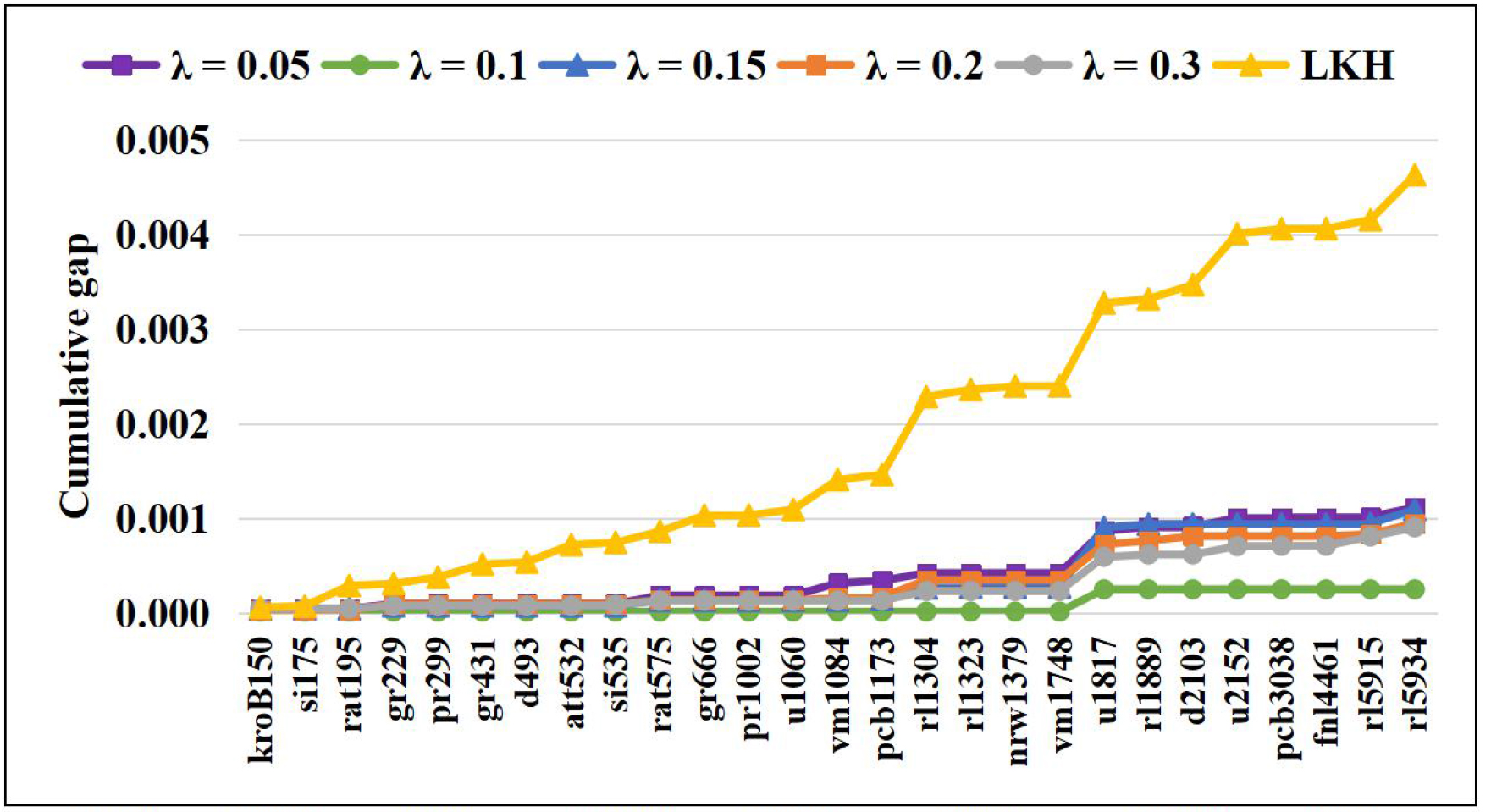} 
\label{fig:pc}
}
\subfigure[The results with different $\gamma$]{
\includegraphics[width=0.3\columnwidth]{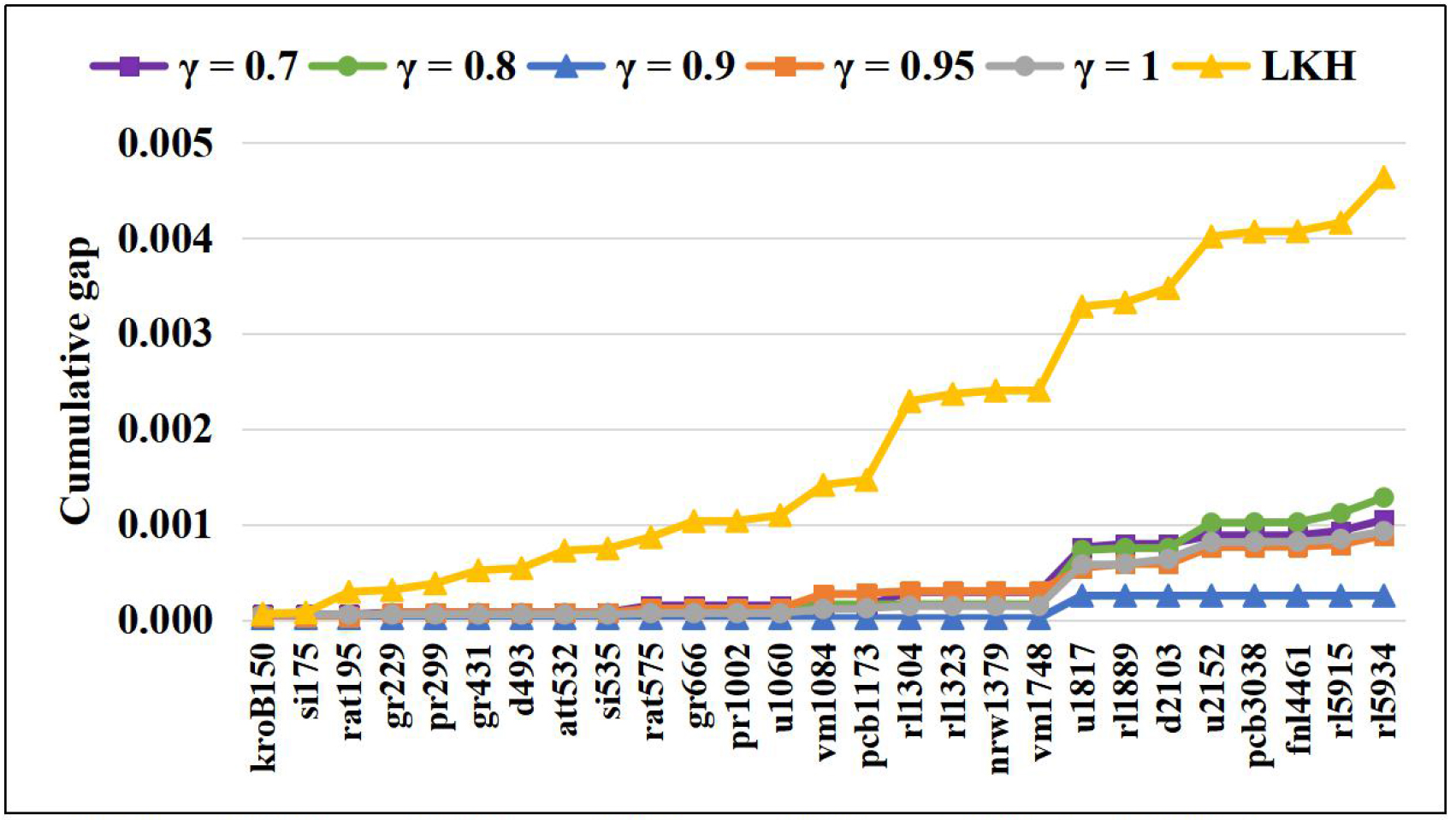} 
\label{fig:pd}
}
\subfigure[The results with different \emph{MaxNum}]{
\includegraphics[width=0.3\columnwidth]{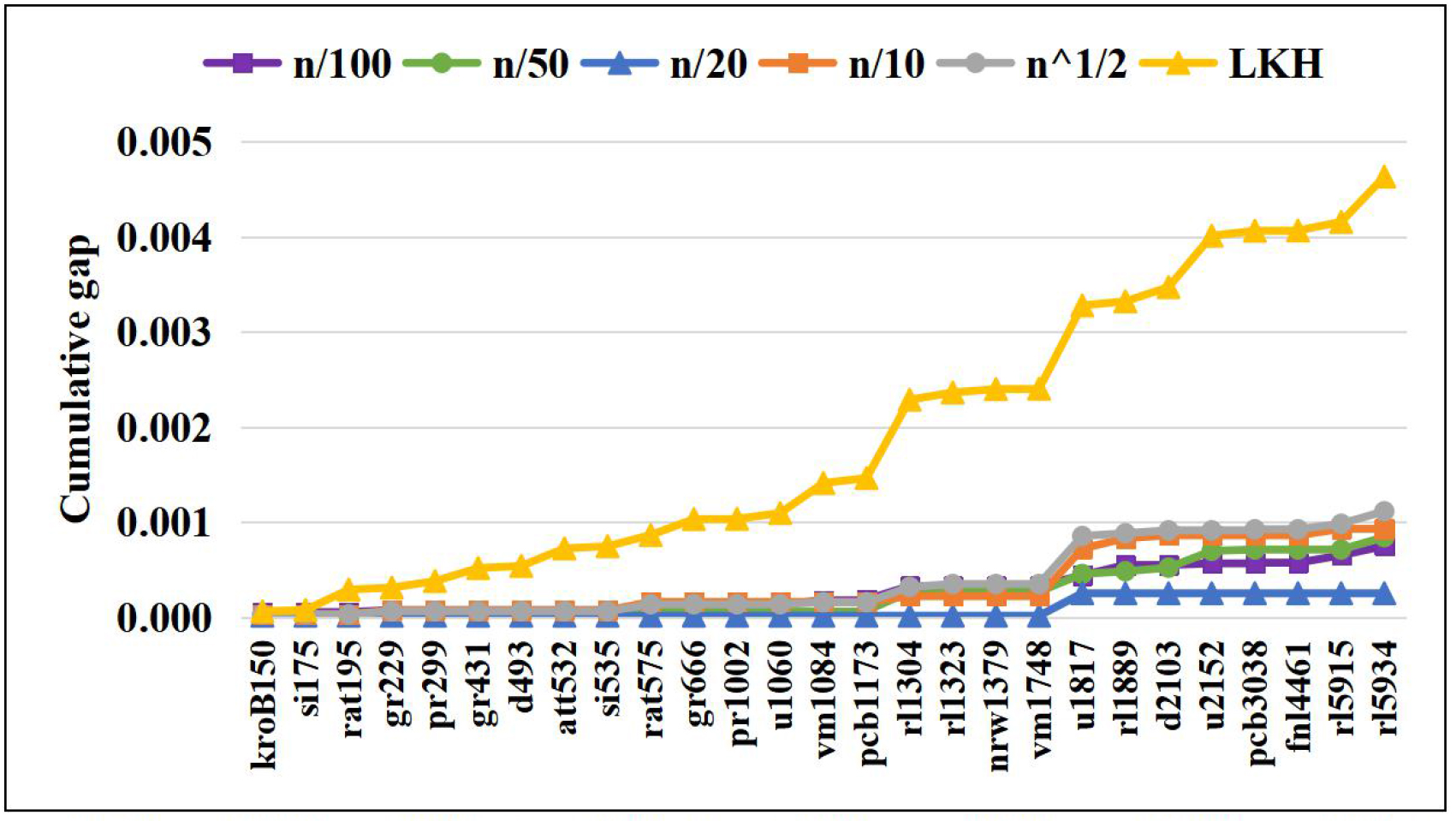} 
\label{fig:pe}
}
\caption{The results of VSR-LKH with different parameters. The 27 \emph{hard} TSP instances are same as the instances in Figure \ref{fig3} and Figure \ref{fig4}. The results are expressed by the cumulative gap (the same in Figure \ref{fig3} and Figure \ref{fig4}).}
\label{fig:parameters}
\end{figure}

The results are shown in Figure \ref{fig:parameters}, and we can observe that the default parameters yield the best results. VSR-LKH with various parameters is significantly better than LKH, indicating that the proposed reinforcement learning method can always improve LKH. In addition, VSR-LKH with various parameters do not show significant difference on the results, indicating that VSR-LKH is robust and not very sensitive to the parameters.

\subsection{Performance Comparison over Time}
To analyze the optimization process of LKH and various variants of VSR-LKH (include VSR-LKH, TD-LKH, Q-LKH, SARSA-LKH, MC-LKH and FixQ-LKH) in solving the TSP, we randomly select six TSP instances (u574, u1060, u1817, fnl4461, rl5934 and rl11849) and compare their results over the runtime. 

\begin{figure}[b]
\centering
\subfigure[The results over time of u574]{
\includegraphics[width=0.3\columnwidth]{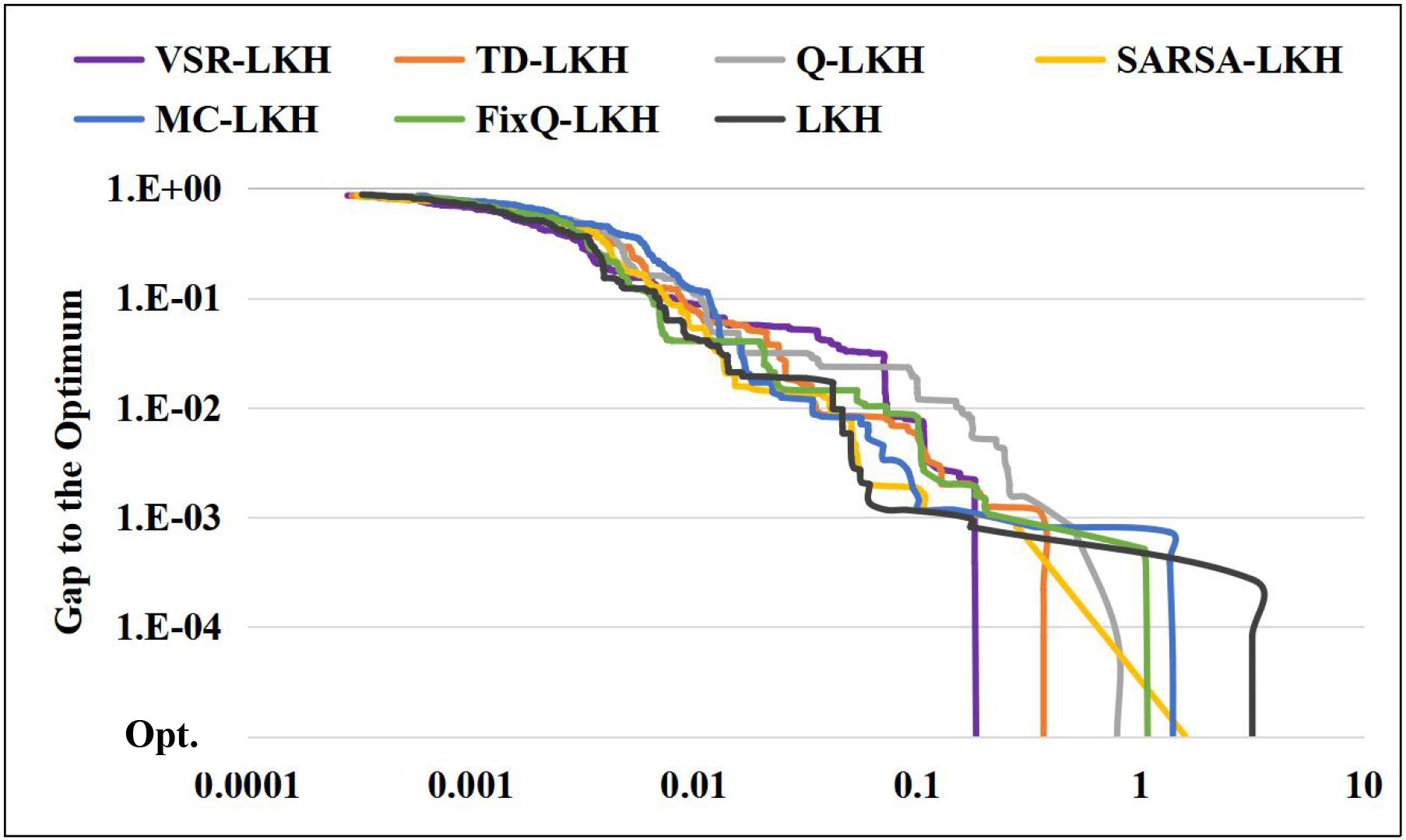} 
\label{fig:ota}
}
\subfigure[The results over time of u1060]{
\includegraphics[width=0.3\columnwidth]{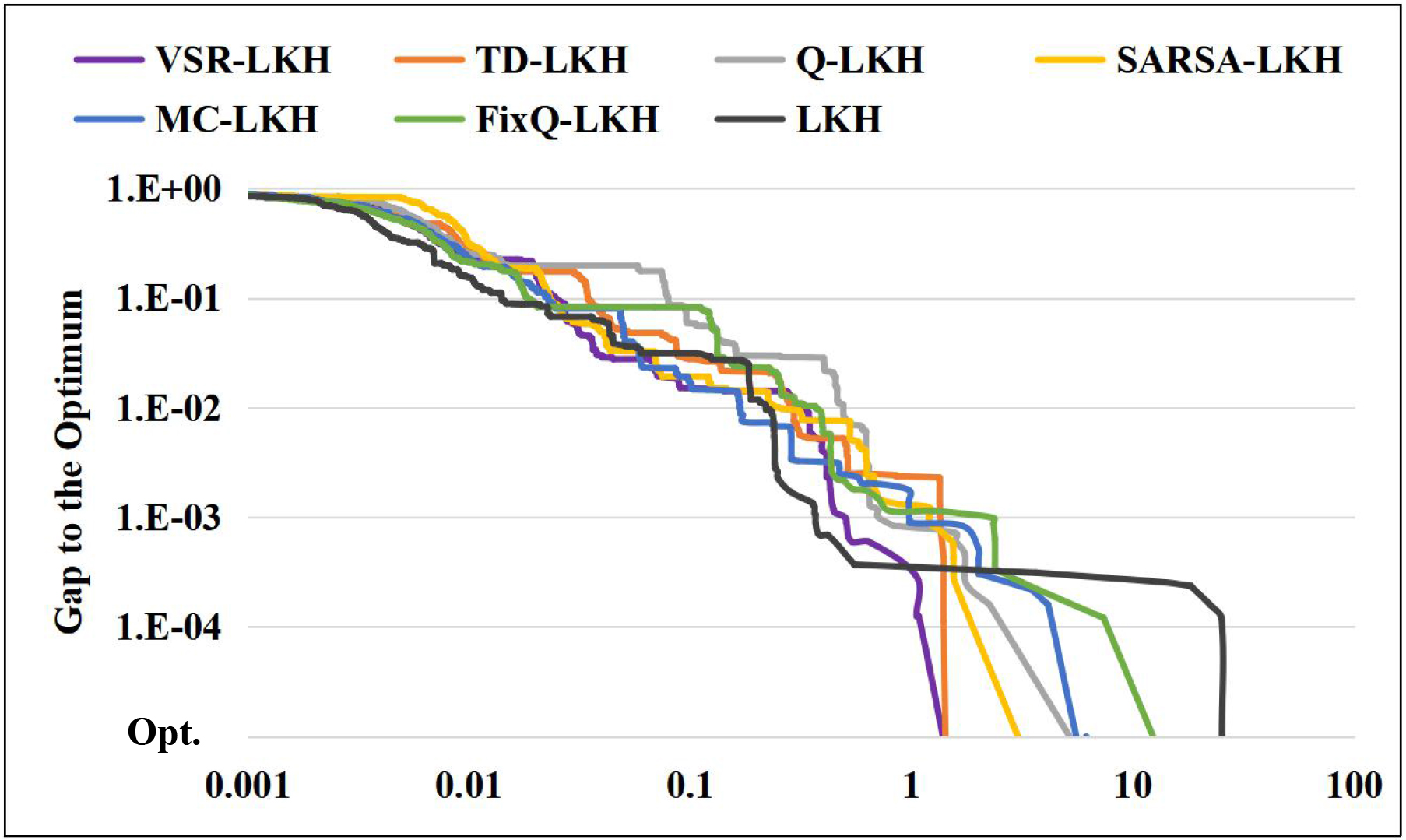} 
\label{fig:otb}
}
\subfigure[The results over time of u1817]{
\includegraphics[width=0.3\columnwidth]{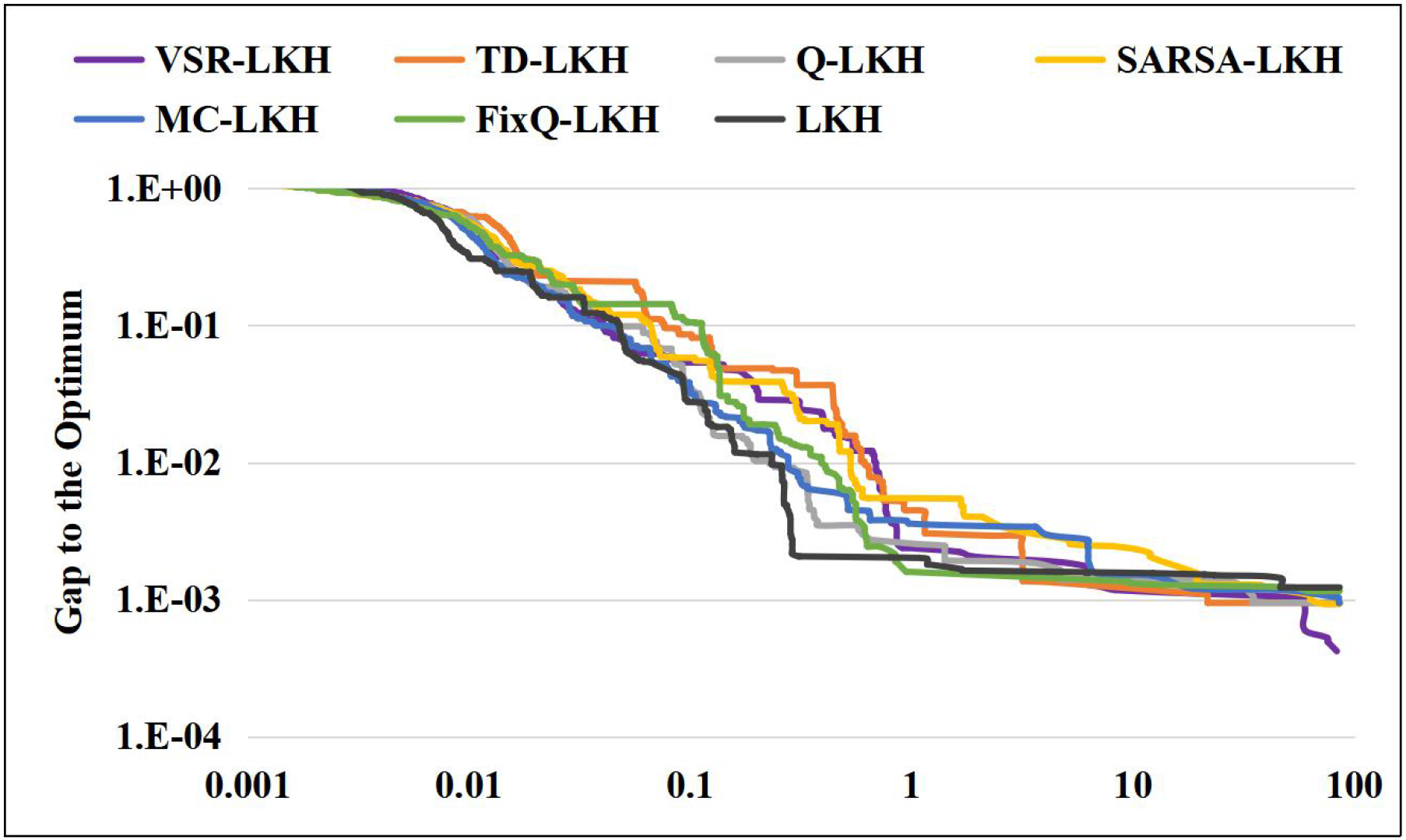} 
\label{fig:otc}
}
\subfigure[The results over time of fnl4461]{
\includegraphics[width=0.3\columnwidth]{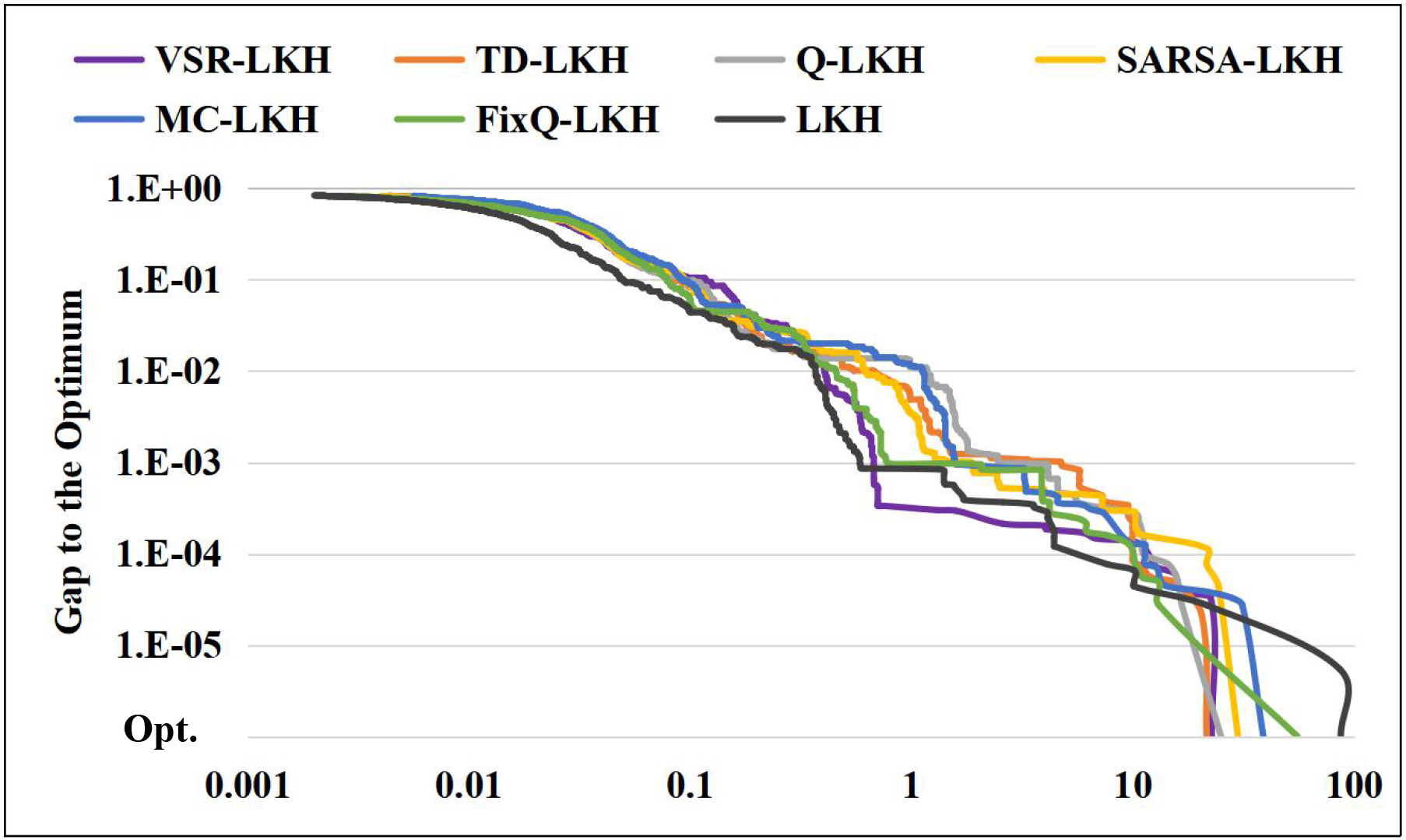} 
\label{fig:otd}
}
\subfigure[The results over time of rl5934]{
\includegraphics[width=0.3\columnwidth]{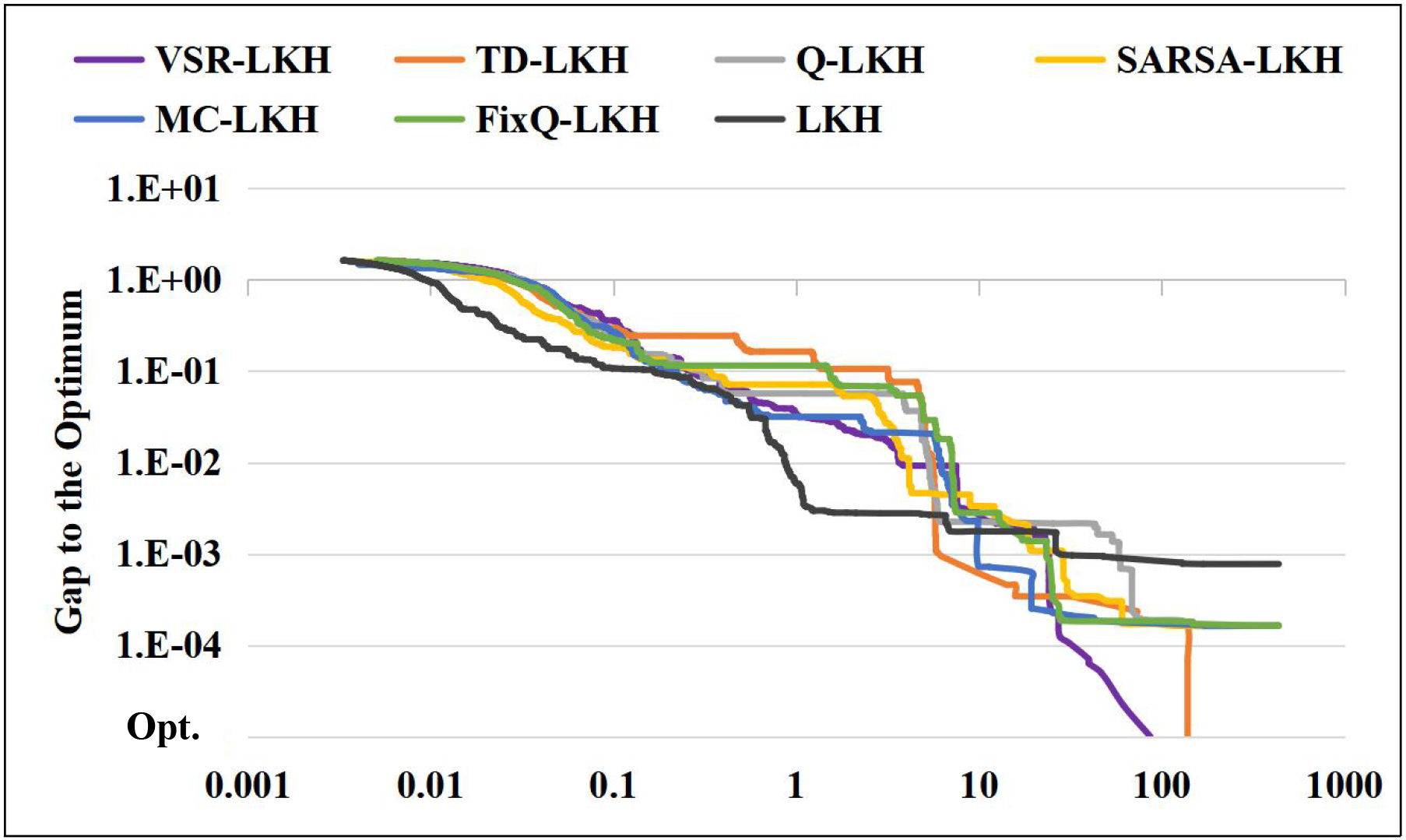} 
\label{fig:ote}
}
\subfigure[The results over time of rl11849]{
\includegraphics[width=0.3\columnwidth]{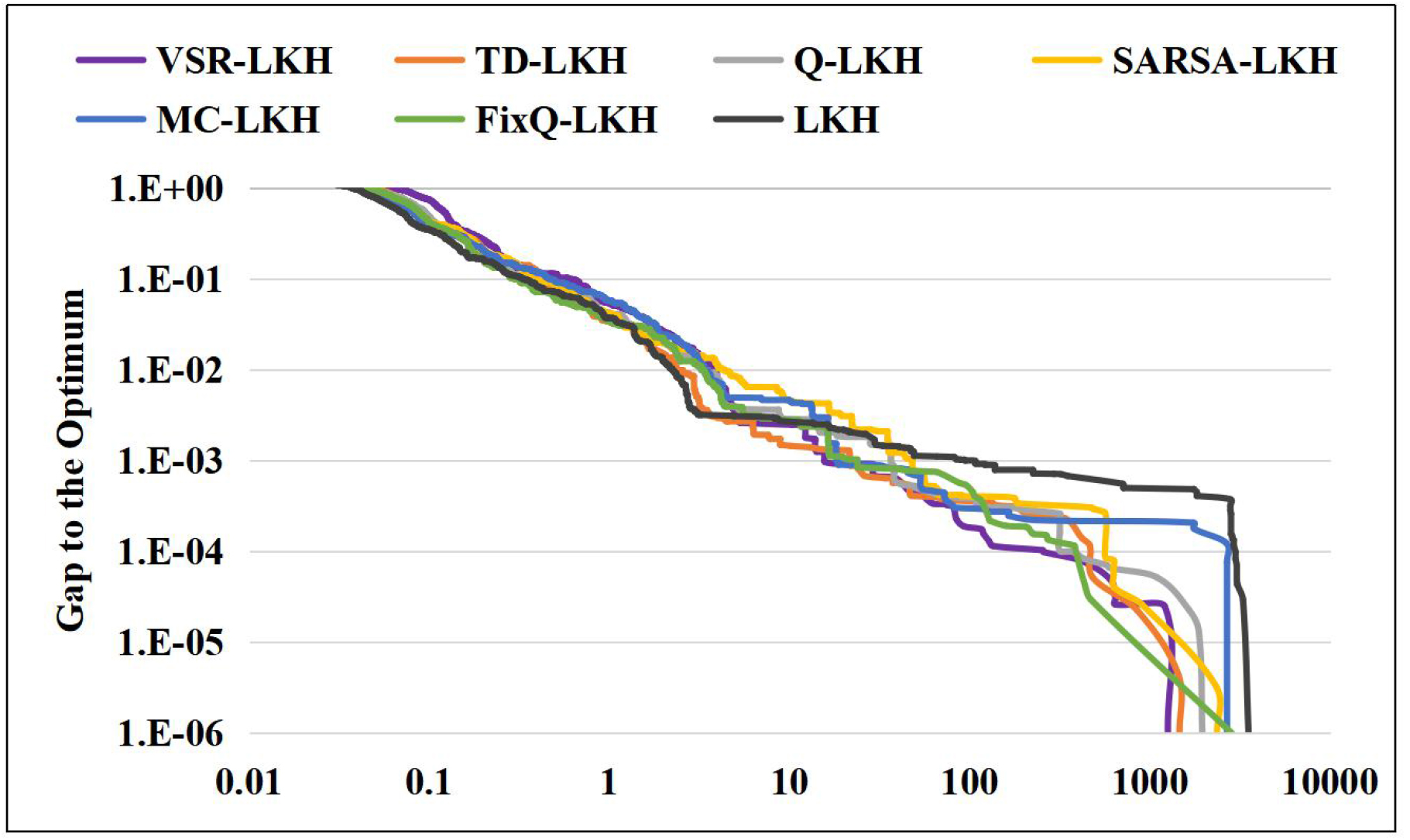} 
\label{fig:otf}
}
\caption{The results over time of LKH and various variants of VSR-LKH. The results are expressed by the gap to the optimal solution of TSP over time in a single run. The horizontal axis is in seconds.}
\label{fig:overtime}
\end{figure}

The results are shown in Figure \ref{fig:overtime}.
In the beginning of the iterations, LKH can yield better solutions than the reinforced algorithms.  This is because the trial and error characteristics of reinforcement learning. However, the reinforced algorithms can surpass LKH in minutes or even seconds when solving these instances, demonstrating that the performance of VSR-LKH variants is better than LKH. In addition, FixQ-LKH also shows better results than LKH in Figure \ref{fig:overtime}, indicating that the performance of LKH can be improved by only replacing the $\alpha$-value with our initial Q-value to select and sort the candidate cities.

\subsection{Full Experimental Results}
We do full comparison of VSR-LKH and LKH on all the 111 symmetric TSP instances from the TSPLIB. 
We run 10 times for almost all instances except two super large instances, pla33810 and
pla85900, due to the resource limit. 
For the two super large instances, we limit the maximum single runtime of LKH and VSR-LKH to 100,000 seconds and only run 3 times for the comparison. 

Table \ref{table3} shows the results. We can see that:
\begin{itemize}
\item
On all the 74 \emph{easy} instances, both VSR-LKH and LKH can easily yield the optimal solution, and there is almost no difference on the running time.
\item On all the 37 \emph{hard} instances, the proposed VSR-LKH greatly promotes the performance of the start-of-the-art algorithm LKH. 
\end{itemize}


\onecolumn

\twocolumn
\onecolumn

\begin{longtable}[h]{lrrrrrrrr}
\caption{Comparison of VSR-LKH and LKH on all 111 instances in TSPLIB (best results in bold).} \label{table3}  \\
\toprule \multicolumn{1}{l}{\textbf{NAME}} & \multicolumn{1}{r}{\textbf{Opt.}} & \multicolumn{1}{r}{\textbf{Method}} & \multicolumn{1}{r}{\textbf{Best}} & \multicolumn{1}{r}{\textbf{Average}} & \multicolumn{1}{r}{\textbf{Worst}} & \multicolumn{1}{r}{\textbf{Success}} & \multicolumn{1}{r}{\textbf{Time(s)}} & \multicolumn{1}{r}{\textbf{Trials}}\\ \hline 
\endfirsthead

\multicolumn{9}{c} {\multirow{2}{*}{\tablename\ \thetable{} -- continued from previous page}} \\ 
\multicolumn{9}{l}{} \\ 
\toprule \multicolumn{1}{l}{\textbf{NAME}} & \multicolumn{1}{r}{\textbf{Opt.}} & \multicolumn{1}{r}{\textbf{Method}} & \multicolumn{1}{r}{\textbf{Best}} & \multicolumn{1}{r}{\textbf{Average}} & \multicolumn{1}{r}{\textbf{Worst}} & \multicolumn{1}{r}{\textbf{Success}} & \multicolumn{1}{r}{\textbf{Time(s)}} & \multicolumn{1}{r}{\textbf{Trials}}\\ \hline 
\endhead


\bottomrule[1pt]
\endlastfoot

\multicolumn{9}{l}{\multirow{2}{*}{\emph{easy} instances (a total of 74)}} \\ 
\multicolumn{9}{l}{} \\ \hline
\multirow{2}{*}{burma14}   & \multirow{2}{*}{3323}     & LKH     & \textbf{Opt.}    & Opt.       & Opt.     & 10/10   & 0.00             & 1.0       \\
                           &                           & VSR-LKH & \textbf{Opt.}    & Opt.       & Opt.     & 10/10   & 0.00             & 1.0       \\ \hline
\multirow{2}{*}{ulysses16} & \multirow{2}{*}{6859}     & LKH     & \textbf{Opt.}    & Opt.       & Opt.     & 10/10   & 0.00             & 1.0       \\
                           &                           & VSR-LKH & \textbf{Opt.}    & Opt.       & Opt.     & 10/10   & 0.00             & 1.0       \\ \hline
\multirow{2}{*}{gr17}      & \multirow{2}{*}{2085}     & LKH     & \textbf{Opt.}    & Opt.       & Opt.     & 10/10   & 0.00             & 1.0       \\
                           &                           & VSR-LKH & \textbf{Opt.}    & Opt.       & Opt.     & 10/10   & 0.00             & 1.0       \\ \hline
\multirow{2}{*}{gr21}      & \multirow{2}{*}{2707}     & LKH     & \textbf{Opt.}    & Opt.       & Opt.     & 10/10   & 0.00             & 1.0       \\
                           &                           & VSR-LKH & \textbf{Opt.}    & Opt.       & Opt.     & 10/10   & 0.00             & 1.0       \\ \hline
\multirow{2}{*}{ulysses22} & \multirow{2}{*}{7013}     & LKH     & \textbf{Opt.}    & Opt.       & Opt.     & 10/10   & 0.00             & 1.0       \\
                           &                           & VSR-LKH & \textbf{Opt.}    & Opt.       & Opt.     & 10/10   & 0.00             & 1.0       \\ \hline
\multirow{2}{*}{gr24}      & \multirow{2}{*}{1272}     & LKH     & \textbf{Opt.}    & Opt.       & Opt.     & 10/10   & 0.00             & 1.0       \\
                           &                           & VSR-LKH & \textbf{Opt.}    & Opt.       & Opt.     & 10/10   & 0.00             & 1.0       \\ \hline
\multirow{2}{*}{fri26}     & \multirow{2}{*}{937}      & LKH     & \textbf{Opt.}    & Opt.       & Opt.     & 10/10   & 0.00             & 1.0       \\
                           &                           & VSR-LKH & \textbf{Opt.}    & Opt.       & Opt.     & 10/10   & 0.00             & 1.0       \\ \hline
\multirow{2}{*}{bayg29}    & \multirow{2}{*}{1610}     & LKH     & \textbf{Opt.}    & Opt.       & Opt.     & 10/10   & 0.00             & 1.0       \\
                           &                           & VSR-LKH & \textbf{Opt.}    & Opt.       & Opt.     & 10/10   & 0.00             & 1.0       \\ \hline
\multirow{2}{*}{bays29}    & \multirow{2}{*}{2020}     & LKH     & \textbf{Opt.}    & Opt.       & Opt.     & 10/10   & 0.00             & 1.0       \\
                           &                           & VSR-LKH & \textbf{Opt.}    & Opt.       & Opt.     & 10/10   & 0.00             & 1.0       \\ \hline
\multirow{2}{*}{dantzig42} & \multirow{2}{*}{699}      & LKH     & \textbf{Opt.}    & Opt.       & Opt.     & 10/10   & 0.00             & 1.0       \\
                           &                           & VSR-LKH & \textbf{Opt.}    & Opt.       & Opt.     & 10/10   & 0.00             & 1.0       \\ \hline
\multirow{2}{*}{swiss42}   & \multirow{2}{*}{1273}     & LKH     & \textbf{Opt.}    & Opt.       & Opt.     & 10/10   & 0.01             & 1.0       \\
                           &                           & VSR-LKH & \textbf{Opt.}    & Opt.       & Opt.     & 10/10   & 0.01             & 1.0       \\ \hline
\multirow{2}{*}{att48}     & \multirow{2}{*}{10628}    & LKH     & \textbf{Opt.}    & Opt.       & Opt.     & 10/10   & 0.01             & 1.0       \\
                           &                           & VSR-LKH & \textbf{Opt.}    & Opt.       & Opt.     & 10/10   & 0.01             & 1.0       \\ \hline
\multirow{2}{*}{gr48}      & \multirow{2}{*}{5046}     & LKH     & \textbf{Opt.}    & Opt.       & Opt.     & 10/10   & 0.02             & 1.0       \\
                           &                           & VSR-LKH & \textbf{Opt.}    & Opt.       & Opt.     & 10/10   & 0.02             & 1.0       \\ \hline
\multirow{2}{*}{hk48}      & \multirow{2}{*}{11461}    & LKH     & \textbf{Opt.}    & Opt.       & Opt.     & 10/10   & 0.01             & 1.0       \\
                           &                           & VSR-LKH & \textbf{Opt.}    & Opt.       & Opt.     & 10/10   & 0.01             & 1.0       \\ \hline
\multirow{2}{*}{eil51}     & \multirow{2}{*}{426}      & LKH     & \textbf{Opt.}    & Opt.       & Opt.     & 10/10   & 0.02             & 1.0       \\
                           &                           & VSR-LKH & \textbf{Opt.}    & Opt.       & Opt.     & 10/10   & 0.02             & 1.0       \\ \hline
\multirow{2}{*}{berlin52}  & \multirow{2}{*}{7542}     & LKH     & \textbf{Opt.}    & Opt.       & Opt.     & 10/10   & 0.01             & 1.0       \\
                           &                           & VSR-LKH & \textbf{Opt.}    & Opt.       & Opt.     & 10/10   & 0.01             & 1.0       \\ \hline
\multirow{2}{*}{brazil58}  & \multirow{2}{*}{25395}    & LKH     & \textbf{Opt.}    & Opt.       & Opt.     & 10/10   & 0.01             & 1.0       \\
                           &                           & VSR-LKH & \textbf{Opt.}    & Opt.       & Opt.     & 10/10   & 0.02             & 1.0       \\ \hline
\multirow{2}{*}{st70}      & \multirow{2}{*}{675}      & LKH     & \textbf{Opt.}    & Opt.       & Opt.     & 10/10   & 0.02             & 1.0       \\
                           &                           & VSR-LKH & \textbf{Opt.}    & Opt.       & Opt.     & 10/10   & 0.02             & 1.0       \\ \hline
\multirow{2}{*}{eil76}     & \multirow{2}{*}{538}      & LKH     & \textbf{Opt.}    & Opt.       & Opt.     & 10/10   & 0.02             & 1.0       \\
                           &                           & VSR-LKH & \textbf{Opt.}    & Opt.       & Opt.     & 10/10   & 0.02             & 1.0       \\ \hline
\multirow{2}{*}{pr76}      & \multirow{2}{*}{108159}   & LKH     & \textbf{Opt.}    & Opt.       & Opt.     & 10/10   & 0.04             & 1.0       \\
                           &                           & VSR-LKH & \textbf{Opt.}    & Opt.       & Opt.     & 10/10   & 0.04             & 1.0       \\ \hline
\multirow{2}{*}{gr96}      & \multirow{2}{*}{55209}    & LKH     & \textbf{Opt.}    & Opt.       & Opt.     & 10/10   & 0.10             & 13.9    \\
                           &                           & VSR-LKH & \textbf{Opt.}    & Opt.       & Opt.     & 10/10   & 0.08             & 14.9    \\ \hline
\multirow{2}{*}{rat99}     & \multirow{2}{*}{1211}     & LKH     & \textbf{Opt.}    & Opt.       & Opt.     & 10/10   & 0.02             & 1.0       \\
                           &                           & VSR-LKH & \textbf{Opt.}    & Opt.       & Opt.     & 10/10   & 0.03             & 1.0       \\ \hline
\multirow{2}{*}{kroA100}   & \multirow{2}{*}{21282}    & LKH     & \textbf{Opt.}    & Opt.       & Opt.     & 10/10   & 0.04             & 1.0       \\
                           &                           & VSR-LKH & \textbf{Opt.}    & Opt.       & Opt.     & 10/10   & 0.04             & 1.0       \\ \hline
\multirow{2}{*}{kroB100}   & \multirow{2}{*}{22141}    & LKH     & \textbf{Opt.}    & Opt.       & Opt.     & 10/10   & 0.05             & 1.2     \\
                           &                           & VSR-LKH & \textbf{Opt.}    & Opt.       & Opt.     & 10/10   & 0.06             & 1.1     \\ \hline
\multirow{2}{*}{kroC100}   & \multirow{2}{*}{20749}    & LKH     & \textbf{Opt.}    & Opt.       & Opt.     & 10/10   & 0.04             & 1.0       \\
                           &                           & VSR-LKH & \textbf{Opt.}    & Opt.       & Opt.     & 10/10   & 0.04             & 1.0       \\ \hline 
\multicolumn{9}{r}{{Continued on next page}}  \\ \hline \\ \\                           
\multirow{2}{*}{kroD100}   & \multirow{2}{*}{21294}    & LKH     & \textbf{Opt.}    & Opt.       & Opt.     & 10/10   & 0.05             & 1.8     \\
                           &                           & VSR-LKH & \textbf{Opt.}    & Opt.       & Opt.     & 10/10   & 0.05             & 1.0       \\ \hline
\multirow{2}{*}{kroE100}   & \multirow{2}{*}{22068}    & LKH     & \textbf{Opt.}    & Opt.       & Opt.     & 10/10   & 0.06             & 3.2     \\
                           &                           & VSR-LKH & \textbf{Opt.}    & Opt.       & Opt.     & 10/10   & 0.06             & 2.9     \\ \hline
\multirow{2}{*}{rd100}     & \multirow{2}{*}{7910}     & LKH     & \textbf{Opt.}    & Opt.       & Opt.     & 10/10   & 0.02             & 1.0       \\
                           &                           & VSR-LKH & \textbf{Opt.}    & Opt.       & Opt.     & 10/10   & 0.03             & 1.0       \\ \hline
\multirow{2}{*}{eil101}    & \multirow{2}{*}{629}      & LKH     & \textbf{Opt.}    & Opt.       & Opt.     & 10/10   & 0.03             & 1.0       \\
                           &                           & VSR-LKH & \textbf{Opt.}    & Opt.       & Opt.     & 10/10   & 0.03             & 1.0       \\ \hline
\multirow{2}{*}{lin105}    & \multirow{2}{*}{14379}    & LKH     & \textbf{Opt.}    & Opt.       & Opt.     & 10/10   & 0.02             & 1.0       \\
                           &                           & VSR-LKH & \textbf{Opt.}    & Opt.       & Opt.     & 10/10   & 0.02             & 1.0       \\ \hline
\multirow{2}{*}{pr107}     & \multirow{2}{*}{44303}    & LKH     & \textbf{Opt.}    & Opt.       & Opt.     & 10/10   & 0.15             & 1.0       \\
                           &                           & VSR-LKH & \textbf{Opt.}    & Opt.       & Opt.     & 10/10   & 0.19             & 1.0       \\ \hline
\multirow{2}{*}{gr120}     & \multirow{2}{*}{6942}     & LKH     & \textbf{Opt.}    & Opt.       & Opt.     & 10/10   & 0.03             & 1.0       \\
                           &                           & VSR-LKH & \textbf{Opt.}    & Opt.       & Opt.     & 10/10   & 0.04             & 1.5     \\ \hline
\multirow{2}{*}{pr124}     & \multirow{2}{*}{59030}    & LKH     & \textbf{Opt.}    & Opt.       & Opt.     & 10/10   & 0.06             & 1.0       \\
                           &                           & VSR-LKH & \textbf{Opt.}    & Opt.       & Opt.     & 10/10   & 0.08             & 1.0       \\ \hline
\multirow{2}{*}{bier127}   & \multirow{2}{*}{118282}   & LKH     & \textbf{Opt.}    & Opt.       & Opt.     & 10/10   & 0.04             & 1.0       \\
                           &                           & VSR-LKH & \textbf{Opt.}    & Opt.       & Opt.     & 10/10   & 0.06             & 1.0       \\ \hline
\multirow{2}{*}{ch130}     & \multirow{2}{*}{6110}     & LKH     & \textbf{Opt.}    & Opt.       & Opt.     & 10/10   & 0.06             & 1.0       \\
                           &                           & VSR-LKH & \textbf{Opt.}    & Opt.       & Opt.     & 10/10   & 0.07             & 1.0       \\ \hline
\multirow{2}{*}{pr136}     & \multirow{2}{*}{96772}    & LKH     & \textbf{Opt.}    & Opt.       & Opt.     & 10/10   & 0.12             & 1.0       \\
                           &                           & VSR-LKH & \textbf{Opt.}    & Opt.       & Opt.     & 10/10   & 0.12             & 1.0       \\ \hline
\multirow{2}{*}{gr137}     & \multirow{2}{*}{69853}    & LKH     & \textbf{Opt.}    & Opt.       & Opt.     & 10/10   & 0.06             & 1.0       \\
                           &                           & VSR-LKH & \textbf{Opt.}    & Opt.       & Opt.     & 10/10   & 0.08             & 1.0       \\ \hline
\multirow{2}{*}{pr144}     & \multirow{2}{*}{58537}    & LKH     & \textbf{Opt.}    & Opt.       & Opt.     & 10/10   & 0.47             & 1.0       \\
                           &                           & VSR-LKH & \textbf{Opt.}    & Opt.       & Opt.     & 10/10   & 0.53             & 1.0       \\ \hline
\multirow{2}{*}{ch150}     & \multirow{2}{*}{6528}     & LKH     & \textbf{Opt.}    & Opt.       & Opt.     & 10/10   & 0.08             & 1.7     \\
                           &                           & VSR-LKH & \textbf{Opt.}    & Opt.       & Opt.     & 10/10   & 0.12             & 13.0      \\ \hline
\multirow{2}{*}{kroA150}   & \multirow{2}{*}{26524}    & LKH     & \textbf{Opt.}     & Opt.        & Opt.      & 10/10   & 0.09             & 3.8     \\
                           &                           & VSR-LKH & \textbf{Opt.}    & Opt.       & Opt.     & 10/10   & 0.07             & 1.0       \\ \hline
\multirow{2}{*}{pr152}     & \multirow{2}{*}{73682}    & LKH     & \textbf{Opt.}     & Opt.        & Opt.      & 10/10   & 0.87             & 29.4    \\
                           &                           & VSR-LKH & \textbf{Opt.}    & Opt.       & Opt.     & 10/10   & 0.68             & 18.3    \\ \hline
\multirow{2}{*}{u159}      & \multirow{2}{*}{42080}    & LKH     & \textbf{Opt.}     & Opt.        & Opt.      & 10/10   & 0.05             & 1.0       \\
                           &                           & VSR-LKH & \textbf{Opt.}    & Opt.       & Opt.     & 10/10   & 0.05             & 1.0       \\ \hline
\multirow{2}{*}{brg180}    & \multirow{2}{*}{1950}     & LKH     & \textbf{Opt.}     & Opt.        & Opt.      & 10/10   & 0.09             & 4.1     \\
                           &                           & VSR-LKH & \textbf{Opt.}    & Opt.       & Opt.     & 10/10   & 0.17             & 6.9     \\ \hline
\multirow{2}{*}{d198}      & \multirow{2}{*}{15780}    & LKH     & \textbf{Opt.}     & Opt.        & Opt.      & 10/10   & 0.71             & 1.0       \\
                           &                           & VSR-LKH & \textbf{Opt.}    & Opt.       & Opt.     & 10/10   & 0.97             & 1.0       \\ \hline
\multirow{2}{*}{kroA200}   & \multirow{2}{*}{29368}    & LKH     & \textbf{Opt.}     & Opt.        & Opt.      & 10/10   & 0.13             & 1.7     \\
                           &                           & VSR-LKH & \textbf{Opt.}    & Opt.       & Opt.     & 10/10   & 0.14             & 1.0       \\ \hline
\multirow{2}{*}{kroB200}   & \multirow{2}{*}{29437}    & LKH     & \textbf{Opt.}     & Opt.        & Opt.      & 10/10   & 0.06             & 1.0       \\
                           &                           & VSR-LKH & \textbf{Opt.}    & Opt.       & Opt.     & 10/10   & 0.08             & 1.0       \\ \hline
\multirow{2}{*}{gr202}     & \multirow{2}{*}{40160}    & LKH     & \textbf{Opt.}     & Opt.        & Opt.      & 10/10   & 0.07             & 1.0       \\
                           &                           & VSR-LKH & \textbf{Opt.}    & Opt.       & Opt.     & 10/10   & 0.10             & 1.4     \\ \hline
\multirow{2}{*}{ts225}     & \multirow{2}{*}{126643}   & LKH     & \textbf{Opt.}     & Opt.        & Opt.      & 10/10   & 0.11             & 1.0       \\
                           &                           & VSR-LKH & \textbf{Opt.}    & Opt.       & Opt.     & 10/10   & 0.11             & 1.0       \\ \hline
\multirow{2}{*}{tsp225}    & \multirow{2}{*}{3916}     & LKH     & \textbf{Opt.}     & Opt.        & Opt.      & 10/10   & 0.12             & 1.0       \\
                           &                           & VSR-LKH & \textbf{Opt.}    & Opt.       & Opt.     & 10/10   & 0.18             & 2.7     \\ \hline
\multirow{2}{*}{pr226}     & \multirow{2}{*}{80369}    & LKH     & \textbf{Opt.}     & Opt.        & Opt.      & 10/10   & 0.14             & 1.0       \\
                           &                           & VSR-LKH & \textbf{Opt.}    & Opt.       & Opt.     & 10/10   & 0.19             & 4.5    \\ \hline
\multirow{2}{*}{gil262}    & \multirow{2}{*}{2378}     & LKH     & \textbf{Opt.}     & Opt.        & Opt.      & 10/10   & 0.23             & 10.6    \\
                           &                           & VSR-LKH & \textbf{Opt.}    & Opt.       & Opt.     & 10/10   & 0.17             & 2.4     \\ \hline
\multicolumn{9}{r}{{Continued on next page}}  \\ \hline \\ \\
\multirow{2}{*}{pr264}     & \multirow{2}{*}{49135}    & LKH     & \textbf{Opt.}     & Opt.        & Opt.      & 10/10   & 0.34             & 14.4    \\
                           &                           & VSR-LKH & \textbf{Opt.}    & Opt.       & Opt.     & 10/10   & 0.30             & 1.4     \\ \hline
\multirow{2}{*}{a280}      & \multirow{2}{*}{2579}     & LKH     & \textbf{Opt.}     & Opt.        & Opt.      & 10/10   & 0.11             & 1.0       \\
                           &                           & VSR-LKH & \textbf{Opt.}    & Opt.       & Opt.     & 10/10   & 0.11             & 1.0       \\ \hline
\multirow{2}{*}{lin318}    & \multirow{2}{*}{42029}    & LKH     & \textbf{Opt.}     & Opt.        & Opt.      & 10/10   & 0.71             & 27.9    \\
                           &                           & VSR-LKH & \textbf{Opt.}    & Opt.       & Opt.     & 10/10   & 0.50             & 14.8    \\ \hline
\multirow{2}{*}{linhp318}  & \multirow{2}{*}{41345}    & LKH     & \textbf{Opt.}     & Opt.        & Opt.      & 10/10   & 0.23             & 11.9    \\
                           &                           & VSR-LKH & \textbf{Opt.}    & Opt.       & Opt.     & 10/10   & 0.27             & 15.1    \\ \hline
\multirow{2}{*}{rd400}     & \multirow{2}{*}{15281}    & LKH     & \textbf{Opt.}     & Opt.        & Opt.      & 10/10   & 0.49             & 33.0      \\
                           &                           & VSR-LKH & \textbf{Opt.}    & Opt.       & Opt.     & 10/10   & 0.30             & 3.2     \\ \hline
\multirow{2}{*}{fl417}     & \multirow{2}{*}{11861}    & LKH     & \textbf{Opt.}     & Opt.        & Opt.      & 10/10   & 5.17             & 7.3     \\
                           &                           & VSR-LKH & \textbf{Opt.}    & Opt.       & Opt.     & 10/10   & 4.31             & 3.9     \\ \hline
\multirow{2}{*}{pr439}     & \multirow{2}{*}{107217}   & LKH     & \textbf{Opt.}     & Opt.        & Opt.      & 10/10   & 1.04             & 39.5    \\
                           &                           & VSR-LKH & \textbf{Opt.}    & Opt.       & Opt.     & 10/10   & 0.50             & 7.3     \\ \hline
\multirow{2}{*}{pcb442}    & \multirow{2}{*}{50778}    & LKH     & \textbf{Opt.}     & Opt.        & Opt.      & 10/10   & 0.37             & 8.2     \\
                           &                           & VSR-LKH & \textbf{Opt.}    & Opt.       & Opt.     & 10/10   & 0.35             & 7.1     \\ \hline
\multirow{2}{*}{ali535}    & \multirow{2}{*}{202339}   & LKH     & \textbf{Opt.}     & Opt.        & Opt.      & 10/10   & 0.70             & 6.6     \\
                           &                           & VSR-LKH & \textbf{Opt.}    & Opt.       & Opt.     & 10/10   & 1.55             & 35.1    \\ \hline
\multirow{2}{*}{pa561}     & \multirow{2}{*}{2763}     & LKH     & \textbf{Opt.}     & Opt.        & Opt.      & 10/10   & 1.06             & 18.3    \\
                           &                           & VSR-LKH & \textbf{Opt.}    & Opt.       & Opt.     & 10/10   & 1.11             & 8.7     \\ \hline
\multirow{2}{*}{u574}      & \multirow{2}{*}{36905}    & LKH     & \textbf{Opt.}     & Opt.        & Opt.      & 10/10   & 1.78             & 149.9   \\
                           &                           & VSR-LKH & \textbf{Opt.}    & Opt.       & Opt.     & 10/10   & 0.66             & 5.5     \\ \hline                           
\multirow{2}{*}{p654}      & \multirow{2}{*}{34643}    & LKH     & \textbf{Opt.}     & Opt.        & Opt.      & 10/10   & 13.73            & 22.9    \\
                           &                           & VSR-LKH & \textbf{Opt.}    & Opt.       & Opt.     & 10/10   & 13.01            & 17.8    \\ \hline
\multirow{2}{*}{d657}      & \multirow{2}{*}{48912}    & LKH     & \textbf{Opt.}     & Opt.        & Opt.      & 10/10   & 1.03             & 33.5    \\
                           &                           & VSR-LKH & \textbf{Opt.}    & Opt.       & Opt.     & 10/10   & 1.22             & 22.2    \\ \hline
\multirow{2}{*}{u724}      & \multirow{2}{*}{41910}    & LKH     & \textbf{Opt.}     & Opt.        & Opt.      & 10/10   & 3.12             & 125.4   \\
                           &                           & VSR-LKH & \textbf{Opt.}    & Opt.       & Opt.     & 10/10   & 1.79             & 18.4    \\ \hline
\multirow{2}{*}{rat783}    & \multirow{2}{*}{8806}     & LKH     & \textbf{Opt.}     & Opt.        & Opt.      & 10/10   & 0.71             & 4.2     \\
                           &                           & VSR-LKH & \textbf{Opt.}    & Opt.       & Opt.     & 10/10   & 0.78             & 3.4     \\ \hline
\multirow{2}{*}{dsj1000}   & \multirow{2}{*}{18660188} & LKH     & \textbf{Opt.}     & Opt.        & Opt.      & 10/10   & 56.10            & 441.4   \\
                           &                           & VSR-LKH & \textbf{Opt.}    & Opt.       & Opt.     & 10/10   & 38.52            & 87.2    \\ \hline
\multirow{2}{*}{si1032}    & \multirow{2}{*}{92650}    & LKH     & \textbf{Opt.}     & Opt.        & Opt.      & 10/10   & 50.19            & 152.0     \\
                           &                           & VSR-LKH & \textbf{Opt.}    & Opt.       & Opt.     & 10/10   & 14.31            & 37.6    \\ \hline
\multirow{2}{*}{d1291}     & \multirow{2}{*}{50801}    & LKH     & \textbf{Opt.}     & Opt.        & Opt.      & 10/10   & 13.68            & 192.1   \\
                           &                           & VSR-LKH & \textbf{Opt.}    & Opt.       & Opt.     & 10/10   & 3.61             & 13.6    \\ \hline
\multirow{2}{*}{u1432}     & \multirow{2}{*}{152970}   & LKH     & \textbf{Opt.}     & Opt.        & Opt.      & 10/10   & 3.03             & 5.3     \\
                           &                           & VSR-LKH & \textbf{Opt.}    & Opt.       & Opt.     & 10/10   & 3.20             & 3.3     \\ \hline
\multirow{2}{*}{d1655}     & \multirow{2}{*}{62128}    & LKH     & \textbf{Opt.}     & Opt.        & Opt.      & 10/10   & 13.98            & 176.0     \\
                           &                           & VSR-LKH & \textbf{Opt.}    & Opt.       & Opt.     & 10/10   & 6.31             & 12.6    \\ \hline
\multirow{2}{*}{u2319}     & \multirow{2}{*}{234256}   & LKH     & \textbf{Opt.}     & Opt.        & Opt.      & 10/10   & 7.12             & 3.1     \\
                           &                           & VSR-LKH & \textbf{Opt.}    & Opt.       & Opt.     & 10/10   & 10.89            & 8.4     \\ \hline
\multirow{2}{*}{pr2392}    & \multirow{2}{*}{378032}   & LKH     & \textbf{Opt.}     & Opt.        & Opt.      & 10/10   & 8.09             & 5.8     \\
                           &                           & VSR-LKH & \textbf{Opt.}    & Opt.       & Opt.     & 10/10   & 9.36             & 8.7     \\ \hline
\multirow{2}{*}{pla7397}   & \multirow{2}{*}{23260728} & LKH     & \textbf{Opt.}     & Opt.        & Opt.      & 10/10   & 375.36           & 632.4   \\
                           &                           & VSR-LKH & \textbf{Opt.}    & Opt.       & Opt.     & 10/10   & 363.31           & 200.0     \\ \hline
\multicolumn{9}{l}{\multirow{2}{*}{\emph{hard} instances (a total of 37)}} \\ 
\multicolumn{9}{l}{} \\ \hline
\multirow{2}{*}{kroB150}   & \multirow{2}{*}{26130}    & LKH     & \textbf{Opt.}     & 26131.6    & 26132    & 2/10    & 0.40             & 128.4   \\
                           &                           & VSR-LKH & \textbf{Opt.}    & 26130.4    & 26132    & 8/10    & 0.34             & 61.5    \\ \hline
\multirow{2}{*}{si175}     & \multirow{2}{*}{21407}    & LKH     & \textbf{Opt.}     & 21407.3    & 21408    & 7/10    & 10.31            & 105.9   \\
                           &                           & VSR-LKH & \textbf{Opt.}    & Opt.       & Opt.     & 10/10   & 3.06             & 33.5    \\ \hline
\multicolumn{9}{r}{{Continued on next page}}  \\ \hline \\ \\
\multirow{2}{*}{rat195}    & \multirow{2}{*}{2323}     & LKH     & \textbf{Opt.}     & 2323.5     & 2328     & 9/10    & 0.28             & 55.0      \\
                           &                           & VSR-LKH & \textbf{Opt.}    & Opt.       & Opt.     & 10/10   & 0.15             & 2.8     \\ \hline
\multirow{2}{*}{gr229}     & \multirow{2}{*}{134602}   & LKH     & \textbf{Opt.}     & 134604.8   & 134616   & 8/10    & 0.57             & 107     \\
                           &                           & VSR-LKH & \textbf{Opt.}    & Opt.       & Opt.     & 10/10   & 0.45             & 57.5    \\ \hline
\multirow{2}{*}{pr299}     & \multirow{2}{*}{48191}    & LKH     & \textbf{Opt.}     & 48194.3    & 48224    & 9/10    & 0.64             & 51.7    \\
                           &                           & VSR-LKH & \textbf{Opt.}    & Opt.       & Opt.     & 10/10   & 0.41             & 6.2     \\ \hline
\multirow{2}{*}{gr431}     & \multirow{2}{*}{171414}   & LKH     & \textbf{Opt.}     & 171437.4   & 171534   & 4/10    & 7.48             & 339.1   \\
                           &                           & VSR-LKH & \textbf{Opt.}    & Opt.       & Opt.     & 10/10   & 4.66             & 76.4    \\ \hline
\multirow{2}{*}{d493}      & \multirow{2}{*}{35002}    & LKH     & \textbf{Opt.}     & 35002.8    & 35004    & 6/10    & 5.24             & 219.6   \\
                           &                           & VSR-LKH & \textbf{Opt.}    & Opt.       & Opt.     & 10/10   & 1.14             & 10.2    \\ \hline
\multirow{2}{*}{att532}    & \multirow{2}{*}{27686}    & LKH     & \textbf{Opt.}     & 27691.1    & 27703    & 7/10    & 4.16             & 238.2   \\
                           &                           & VSR-LKH & \textbf{Opt.}    & Opt.       & Opt.     & 10/10   & 1.18             & 17.8    \\ \hline
\multirow{2}{*}{si535}     & \multirow{2}{*}{48450}    & LKH     & \textbf{Opt.}     & 48451.1    & 48455    & 7/10    & 36.93            & 311.6   \\
                           &                           & VSR-LKH & \textbf{Opt.}    & Opt.       & Opt.     & 10/10   & 27.62            & 144.0     \\ \hline
\multirow{2}{*}{rat575}    & \multirow{2}{*}{6773}     & LKH     & \textbf{Opt.}     & 6773.8     & 6774     & 2/10    & 3.73             & 526.9   \\
                           &                           & VSR-LKH & \textbf{Opt.}    & Opt.       & Opt.     & 10/10   & 3.69             & 151.3   \\ \hline
\multirow{2}{*}{gr666}     & \multirow{2}{*}{294358}   & LKH     & \textbf{Opt.}     & 294407.4   & 294476   & 5/10    & 6.67             & 467.5   \\
                           &                           & VSR-LKH & \textbf{Opt.}    & Opt.       & Opt.     & 10/10   & 3.33             & 45.2    \\ \hline
\multirow{2}{*}{pr1002}    & \multirow{2}{*}{259045}   & LKH     & \textbf{Opt.}     & 259045.6   & 259048   & 8/10    & 5.98             & 549.0     \\
                           &                           & VSR-LKH & \textbf{Opt.}    & Opt.       & Opt.     & 10/10   & 2.25             & 22.1    \\ \hline
\multirow{2}{*}{u1060}     & \multirow{2}{*}{224094}   & LKH     & \textbf{Opt.}     & 224107.5   & 224121   & 5/10    & 142.07           & 663.3   \\
                           &                           & VSR-LKH & \textbf{Opt.}    & Opt.       & Opt.     & 10/10   & 5.68             & 13.7    \\ \hline
\multirow{2}{*}{vm1084}    & \multirow{2}{*}{239297}   & LKH     & \textbf{Opt.}     & 239372.6   & 239432   & 3/10    & 50.19            & 824.1   \\
                           &                           & VSR-LKH & \textbf{Opt.}    & Opt.       & Opt.     & 10/10   & 12.71            & 116.6   \\ \hline
\multirow{2}{*}{pcb1173}   & \multirow{2}{*}{56892}    & LKH     & \textbf{Opt.}     & 56895.0      & 56897    & 4/10    & 7.15             & 844.0     \\
                           &                           & VSR-LKH & \textbf{Opt.}    & Opt.       & Opt.     & 10/10   & 7.80             & 291.1   \\ \hline
\multirow{2}{*}{rl1304}    & \multirow{2}{*}{252948}   & LKH     & \textbf{Opt.}     & 253156.4   & 253354   & 3/10    & 22.37            & 1170.0    \\
                           &                           & VSR-LKH & \textbf{Opt.}    & Opt.       & Opt.     & 10/10   & 10.44            & 173.0     \\ \hline
\multirow{2}{*}{rl1323}    & \multirow{2}{*}{270199}   & LKH     & \textbf{Opt.}     & 270219.6   & 270324   & 6/10    & 17.55            & 718.8   \\
                           &                           & VSR-LKH & \textbf{Opt.}    & Opt.       & Opt.     & 10/10   & 11.38            & 111.6   \\ \hline
\multirow{2}{*}{nrw1379}   & \multirow{2}{*}{56638}    & LKH     & \textbf{Opt.}     & 56640.0      & 56643    & 6/10    & 15.91            & 759.3   \\
                           &                           & VSR-LKH & \textbf{Opt.}    & Opt.       & Opt.     & 10/10   & 15.84            & 198.2   \\ \hline
\multirow{2}{*}{fl1400}    & \multirow{2}{*}{20127}    & LKH     & \textbf{Opt.}    & 20160.3    & 20164    & 1/10    & 5284.60          & 1372.9  \\
                           &                           & VSR-LKH & \textbf{Opt.}    & Opt.       & Opt.     & 10/10   & 745.20           & 137.2   \\ \hline
\multirow{2}{*}{fl1577}    & \multirow{2}{*}{22249}    & LKH     & \textbf{22254}   & 22260.6    & 22263    & 0/10    & 2541.50          & 1577.0    \\
                           &                           & VSR-LKH & \textbf{22254}   & 22254.0      & 22254    & 0/10    & 10770.90         & 1577.0  \\ \hline
\multirow{2}{*}{vm1748}    & \multirow{2}{*}{336556}   & LKH     & \textbf{Opt.}     & 336557.3   & 336569   & 9/10    & 22.82            & 1007.9  \\
                           &                           & VSR-LKH & \textbf{Opt.}    & Opt.       & Opt.     & 10/10   & 14.08            & 47.0      \\ \hline
\multirow{2}{*}{u1817}     & \multirow{2}{*}{57201}    & LKH     & \textbf{Opt.}     & 57251.1    & 57274    & 1/10    & 116.39           & 1817.0    \\
                           &                           & VSR-LKH & \textbf{Opt.}    & 57214.5    & 57254    & 7/10    & 256.79           & 766.9   \\ \hline
\multirow{2}{*}{rl1889}    & \multirow{2}{*}{316536}   & LKH     & 316549  & 316549.8   & 316553   & 0/10    & 109.78           & 1889.0    \\
                           &                           & VSR-LKH & \textbf{Opt.}    & Opt.       & Opt.     & 10/10   & 26.30            & 91.9    \\ \hline
\multirow{2}{*}{d2103}     & \multirow{2}{*}{80450}    & LKH     & 80454   & 80462.0      & 80473    & 0/10    & 216.48           & 2103.0    \\
                           &                           & VSR-LKH & \textbf{Opt.}    & Opt.       & Opt.     & 10/10   & 121.97           & 511.8   \\ \hline
\multirow{2}{*}{u2152}     & \multirow{2}{*}{64253}    & LKH     & \textbf{Opt.}     & 64287.7    & 64310    & 3/10    & 124.58           & 1614.0    \\
                           &                           & VSR-LKH & \textbf{Opt.}    & Opt.       & Opt.     & 10/10   & 65.41            & 185.8   \\ \hline
\multirow{2}{*}{pcb3038}   & \multirow{2}{*}{137694}   & LKH     & \textbf{Opt.}     & 137701.2   & 137741   & 4/10    & 128.20           & 2078.6  \\
                           &                           & VSR-LKH & \textbf{Opt.}    & Opt.       & Opt.     & 10/10   & 146.11           & 389.3   \\ \hline
\multirow{2}{*}{fl3795}    & \multirow{2}{*}{28772}    & LKH     & 28813   & 28813.7    & 28815    & 0/10    & 54754.9          & 3795.0    \\
                           &                           & VSR-LKH & \textbf{Opt.}    & Opt.       & Opt.     & 10/10   & 1805.20          & 89.4    \\ \hline
\multirow{2}{*}{fnl4461}   & \multirow{2}{*}{182566}   & LKH     & \textbf{Opt.}     & 182566.5   & 182571   & 9/10    & 79.40            & 923.1   \\
                           &                           & VSR-LKH & \textbf{Opt.}    & Opt.       & Opt.     & 10/10   & 71.94            & 94.8    \\ \hline
\multicolumn{9}{r}{{Continued on next page}}  \\ \hline \\ \\
\multirow{2}{*}{rl5915}    & \multirow{2}{*}{565530}   & LKH     & 565544  & 565581.2   & 565593   & 0/10    & 494.81           & 5915.0    \\
                           &                           & VSR-LKH & \textbf{Opt.}    & Opt.       & Opt.     & 10/10   & 438.50           & 851.4   \\ \hline
\multirow{2}{*}{rl5934}    & \multirow{2}{*}{556045}   & LKH     & 556136  & 556309.8   & 556547   & 0/10    & 753.22           & 5934.0    \\
                           &                           & VSR-LKH & \textbf{Opt.}    & Opt.       & Opt.     & 10/10   & 118.78           & 144.6   \\ \hline
\multirow{2}{*}{rl11849}   & \multirow{2}{*}{923288}   & LKH     & \textbf{Opt.}    & 923362.7   & 923532   & 2/10    & 3719.35          & 10933.4 \\
                           &                           & VSR-LKH & \textbf{Opt.}    & Opt.       & Opt.     & 10/10   & 1001.11          & 751.9   \\ \hline
\multirow{2}{*}{usa13509}  & \multirow{2}{*}{19982859} & LKH     & \textbf{Opt.}    & 19983103.4 & 19983569 & 1/10    & 4963.52          & 13509.0   \\
                           &                           & VSR-LKH & \textbf{Opt.}    & 19982930.2 & 19983029 & 5/10    & 25147.63         & 11900.0   \\ \hline
\multirow{2}{*}{brd14051}  & \multirow{2}{*}{469385}   & LKH     & 469393  & 469398.3   & 469420   & 0/10    & 6957.07          & 14051.0   \\
                           &                           & VSR-LKH & \textbf{Opt.}    & 469393.8   & 469403   & 1/10    & 37954.66          & 13581.3 \\ \hline
\multirow{2}{*}{d15112}    & \multirow{2}{*}{1573084}  & LKH     & 1573085 & 1573142.7  & 1573223  & 0/10    & 7287.69          & 15112.0   \\
                           &                           & VSR-LKH & \textbf{Opt.}    & 1573125.9  & 1573215  & 2/10    & 45642.47          & 13185.3  \\ \hline
\multirow{2}{*}{d18512}    & \multirow{2}{*}{645238}   & LKH     & \textbf{645239}  & 645260.6   & 645286   & 0/10    & 12399.27         & 18512.0  \\
                           &                           & VSR-LKH & 645241  & 645261.6   & 645273   & 0/10    & 62662.90         & 18512.0   \\ \hline
\multirow{2}{*}{pla33810}  & \multirow{2}{*}{66048945} & LKH     & 66061689 & 66062670.0  & 66064532  & 0/3    & 100028.76          & 4294.3   \\
                           &                           & VSR-LKH & \textbf{66049981}  & 66055810.7  & 66063427  & 0/3    & 100028.91           & 1724.7  \\ \hline
\multirow{2}{*}{pla85900}  & \multirow{2}{*}{142382641} & LKH     & 142418173 & 142422390.7  & 142424554  & 0/3    & 100004.28          & 16486.3   \\
                           &                           & VSR-LKH  & \textbf{142393330} & 142396406.7  & 142402355  & 0/3    & 100018.84          & 2895.3  \\
\end{longtable}

\twocolumn

\end{document}